\documentclass[]{fairmeta} 

\title{How much do language models memorize?}

\author[1,3]{John X. Morris}
\author[2]{Chawin Sitawarin}
\author[1]{Chuan Guo}
\author[1]{Narine Kokhlikyan}

\author[3,4]{G. Edward Suh}
\author[3]{\\ Alexander M. Rush}
\author[1]{Kamalika Chaudhuri}
\author[1]{Saeed Mahloujifar}

\affiliation[1]{FAIR at Meta}
\affiliation[2]{Google DeepMind}
\affiliation[3]{Cornell University}
\affiliation[4]{NVIDIA}

\date{\today}
\correspondence{Saeed Mahloujifar at \email{saeedm@meta.com}}
\usepackage{caption}

\usepackage{microtype}
\usepackage{graphicx}
\usepackage{hyperref}

\usepackage[textsize=tiny]{todonotes}
\usepackage{microtype}
\usepackage{graphicx}
\usepackage{booktabs} 

\usepackage{amsfonts}
\usepackage{amsmath}
\usepackage{amssymb}
\usepackage{amsthm}
\usepackage{mathtools}
\usepackage{multicol}
\usepackage{multirow}
\usepackage{framed}
\usepackage{CJKutf8}

\theoremstyle{plain}
\theoremstyle{definition}
\theoremstyle{remark}

\newcounter{thm}

\newtheorem{lemma}[thm]{Lemma}

\newtheorem{proposition}[thm]{Proposition}

\newtheorem{definition}[thm]{Definition}

\DeclareMathOperator*{\E}{E}
\newcommand{\Ex}{\E}

\newcommand{\set}[1]{\{#1\}}

\usepackage{etoolbox}

\newcommand{\definecalletter}[1]{%
  \expandafter\newcommand\csname c#1\endcsname{\mathcal{#1}}%
}
\forcsvlist{\definecalletter}{A,B,C,D,E,F,G,H,I,J,K,L,M,N,O,P,Q,R,S,T,U,V,W,X,Y,Z}

\DeclarePairedDelimiterX{\infdivx}[2]{(}{)}{%
  #1\;\delimsize\|\;#2%
}

\usepackage{xcolor}

\usepackage{amsmath}

\newcommand{\unintendedMemorizationName}{unintended memorization}
\newcommand{\unintendedMemorizationNameCapital}{Unintended memorization}

\newcommand{\intendedMemorizationName}{generalization}

\abstract{
We propose a new method for estimating how much a model ``knows'' about a datapoint and use it to measure the capacity of modern language models. We formally separate memorization into two components: \textit{unintended memorization}, the information a model contains about a specific dataset, and \textit{generalization}, the information a model contains about the true data-generation process. By eliminating generalization, we can compute the total memorization of a given model, which provides an estimate of model capacity: our measurements estimate that \textbf{models in the GPT family have an approximate capacity of 3.6 bits-per-parameter}. We train language models on datasets of increasing size and observe that models memorize until their capacity fills, at which point ``grokking'' begins, and unintended memorization decreases as models begin to generalize. We train hundreds of transformer language models ranging from $500K$ to $1.5B$ parameters and produce a series of scaling laws relating model capacity and data size to membership inference.}

\begin{document}

\maketitle

\begin{figure*}[h!]
    \centering
    \begin{minipage}{0.47\textwidth}
        \centering
        \includegraphics[width=\textwidth]{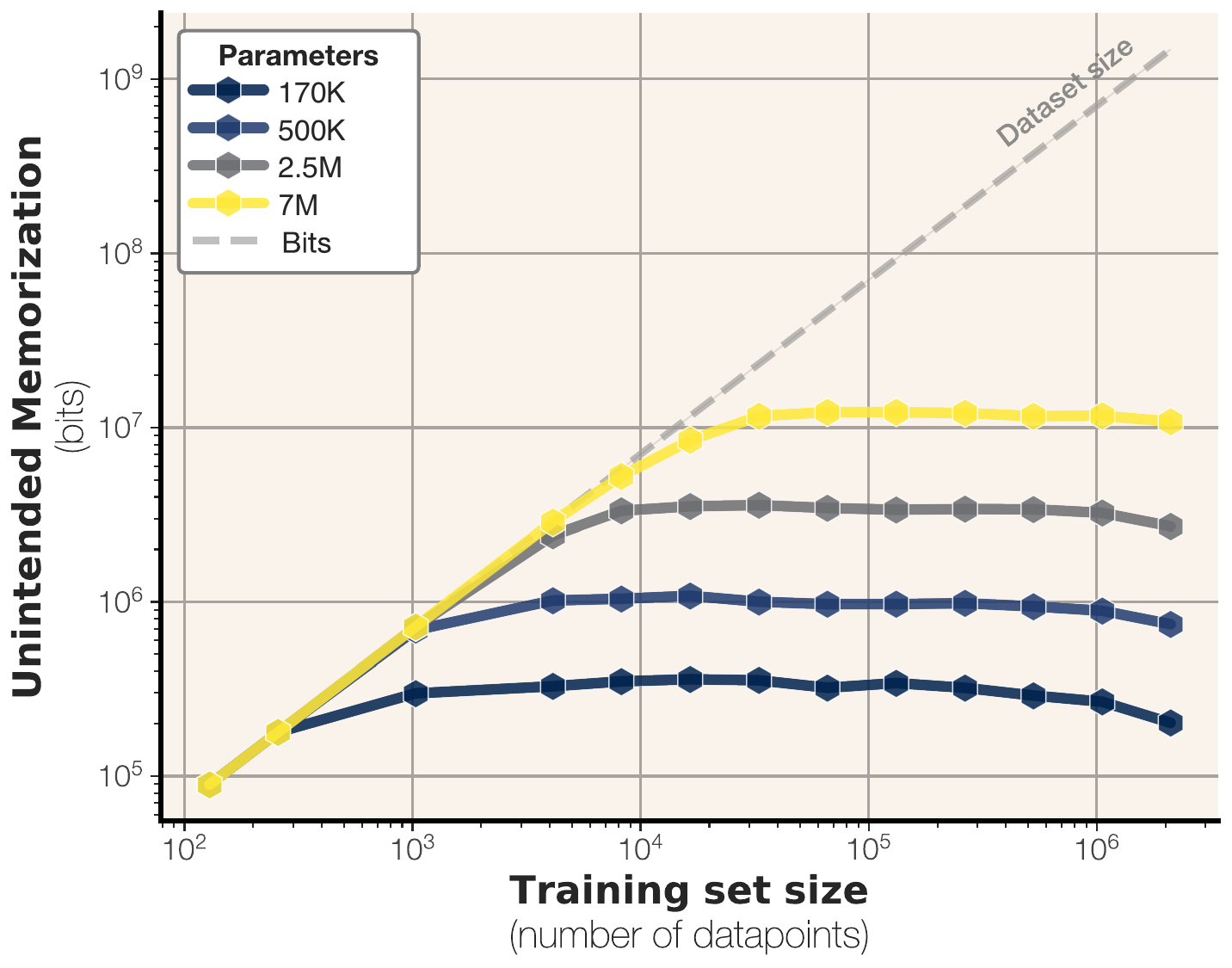}
        \caption{ \textbf{\unintendedMemorizationNameCapital{} of uniform random data} (Section \ref{sec:um-synthetic}). Memorization plateaus at the empirical capacity limit of different-sized models from the GPT-family, approximately 3.6 bits-per-parameter. 
        }
        \label{fig:model_capacity_synthetic}
    \end{minipage}
    \hfill
    \begin{minipage}{0.47\textwidth}
        \centering
        \includegraphics[width=\textwidth]{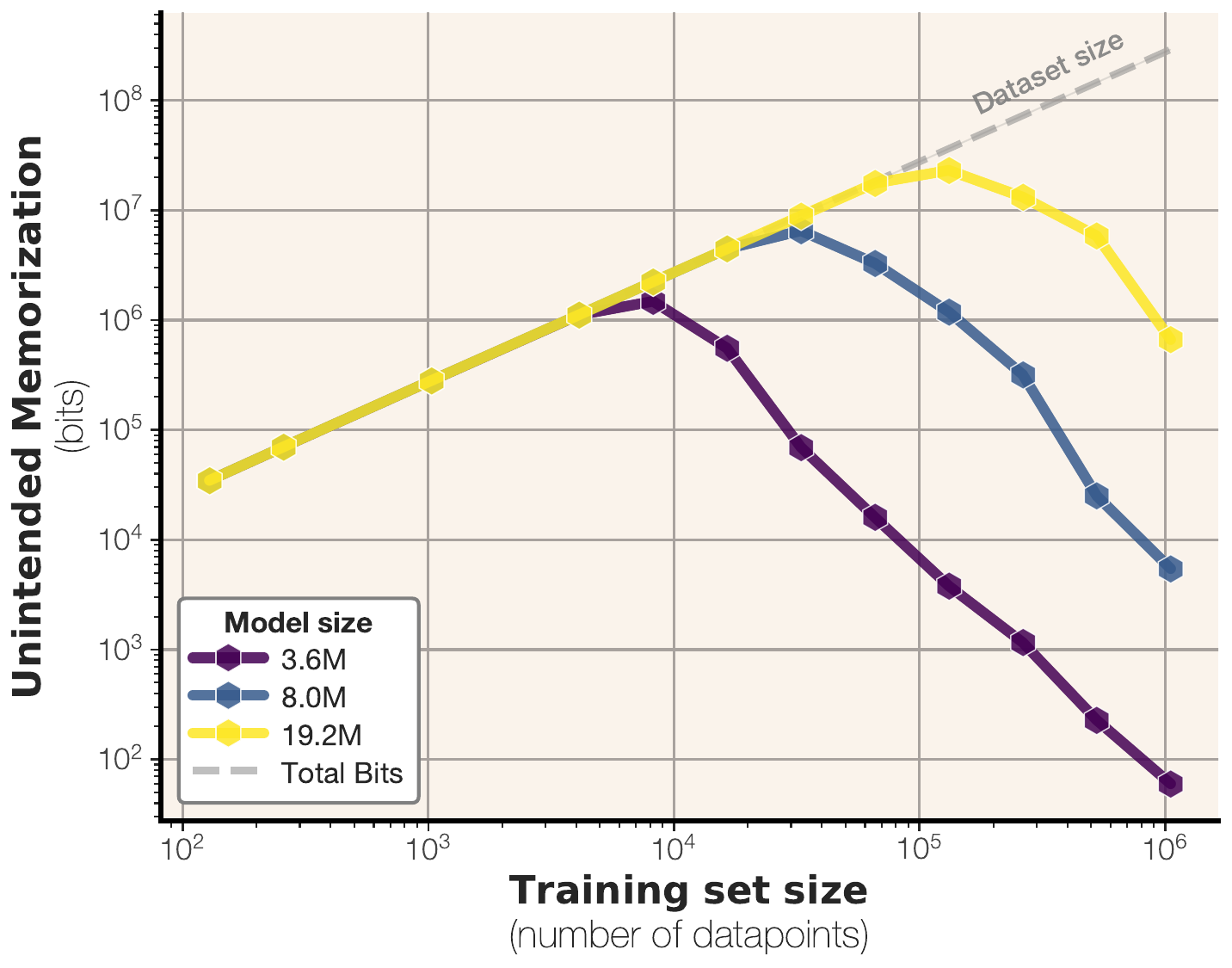}
        \caption{\textbf{\unintendedMemorizationNameCapital{} of text} across model and dataset sizes (Section \ref{sec:um-text}). All quantities are calculated with respect to a large oracle model trained on the full data distribution.
        }
        \label{fig:text-kolmogorov-total-oracle}
    \end{minipage}
\end{figure*}

\section{Introduction}

\begin{figure}[t!]
   \centering
   \begin{minipage}[t]{0.47\textwidth}
       \centering
       \includegraphics[width=\textwidth]{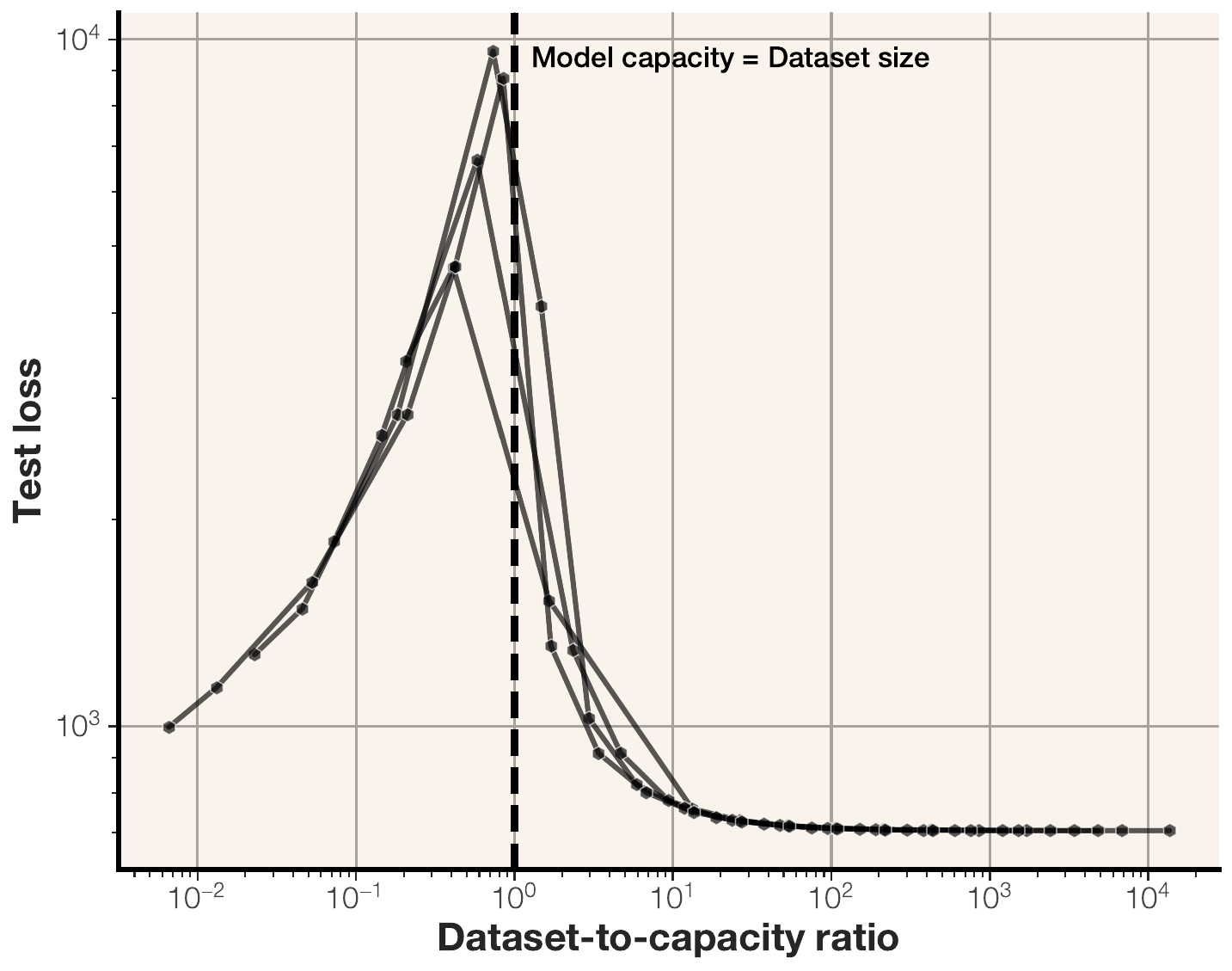}
       \caption{In our experiments on synthetic bitstrings, double descent occurs exactly when the dataset size begins to exceed the model's capacity, when \unintendedMemorizationName{}  is no longer beneficial for lowering the loss.}
       \label{fig:synth_double_descent}
   \end{minipage}
   \hfill
   \begin{minipage}[t]{0.47\textwidth}
       \centering
       \includegraphics[width=\textwidth]{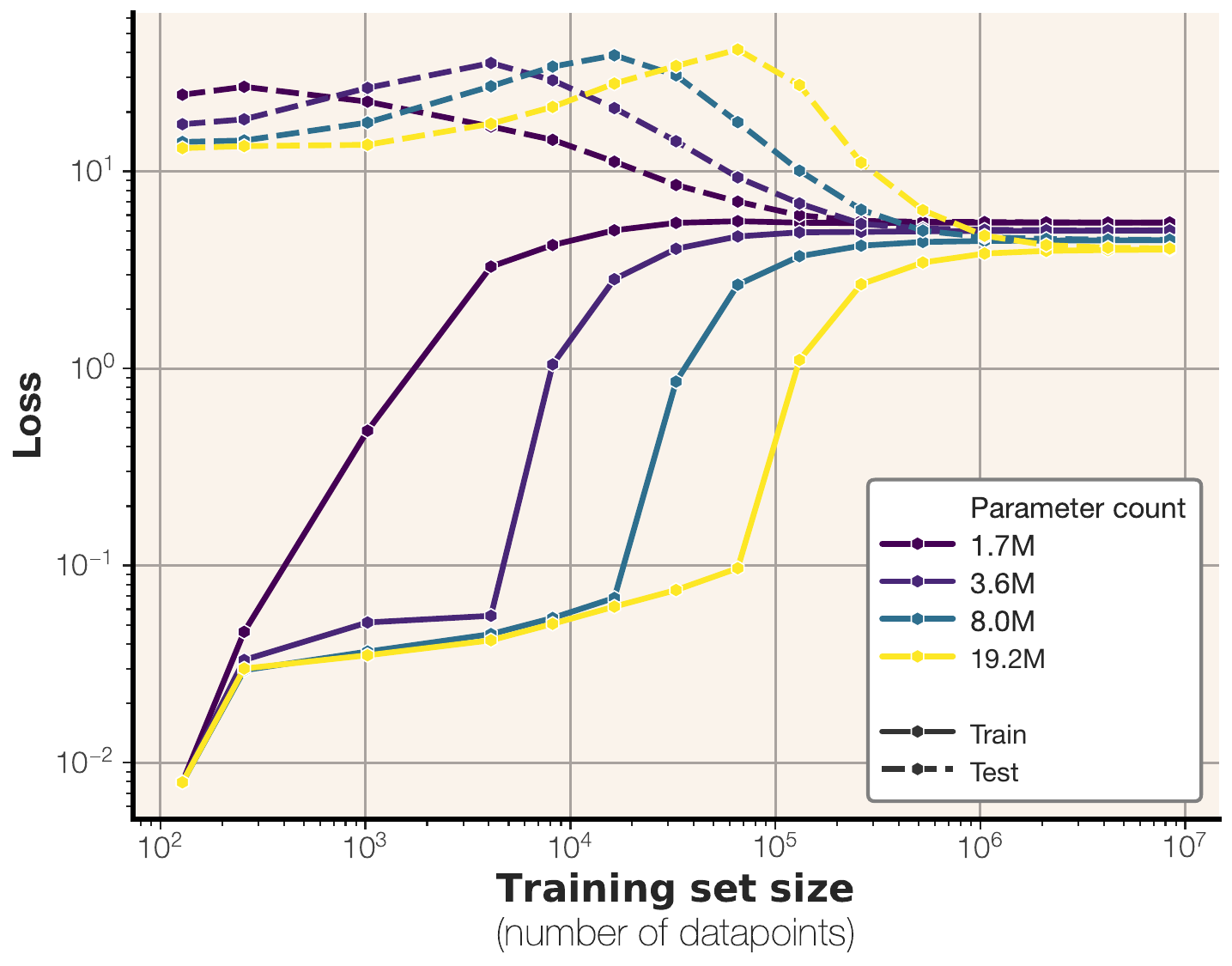}
       \caption{Train and test losses of different model and dataset sizes trained on text. Double descent occurs when dataset size exceeds model capacity.}
       \label{fig:text-loss-train-val}
   \end{minipage}
\end{figure}

For the past several years, modern language models have been trained on increasingly large amounts of data, while parameter counts stay stagnant in the billions. For example, one recent state-of-the-art model \citep{dubey2024llama3herdmodels} has 8 billion parameters (around $32 GB$ on disk) but is trained on 15 trillion tokens (around $7 TB$ on disk).

A long line of work \citep{carlini2019secretsharerevaluatingtesting,mireshghallah2022memorizationnlpfinetuningmethods,nasr2023scalableextractiontrainingdata,zhang2023counterfactualmemorizationneurallanguage,carlini2023quantifyingmemorizationneurallanguage,schwarzschild2024rethinkingllmmemorizationlens} questions whether such pretrained language models memorize their training data in a meaningful way. Most research approaches this problem either through the lens of extraction, aiming to recover full training datapoints from model weights, or membership inference, classifying whether a training point was present in the training data of a given model.

Studies of language model extraction argue that a datapoint is memorized if we can induce the model to generate it \citep{carlini2023quantifyingmemorizationneurallanguage, nasr2023scalableextractiontrainingdata, schwarzschild2024rethinkingllmmemorizationlens}. We argue that such generation does not necessarily serve as a proof of memorization. Language models can be coerced to output almost any string \citep{geiping2024coercingllmsrevealalmost}; hence the fact that a model outputs something is not necessarily a sign of memorization. To address this issue, some researchers have suggested regularizing the input to the language model, such as by limiting its length \citep{schwarzschild2024rethinkingllmmemorizationlens} or matching it to the prefix \citep{carlini2023quantifyingmemorizationneurallanguage} preceding the memorized sentence. However, none of these constraints allow us to distinguish whether a model outputs a string due to memorization or good generalization.  For example, a language model prompted to add two numbers can output the answer without having seen the equation before.


 
To address this issue, we propose a definition of memorization that quantifies the extent to which a model retains information about a specific datapoint. Our approach leverages the concept of compression rate in bits: a model is considered to have memorized an input if the input can be compressed into a shorter encoding when the model is available. This framework draws inspiration from Kolmogorov information theory \citep{kolmogorov1963tables} and Shannon information theory \citep{shannon1948noisychannel}, but remains practical by estimating information content using model likelihoods. We address the fundamental challenge of distinguishing memorization from generalization \citep{prashanth2024recitereconstructrecollect} by decomposing memorization into two distinct components: \textit{\unintendedMemorizationName}, which captures the information the model retains about a specific dataset, and \textit{\intendedMemorizationName}, which represents the knowledge the model acquires about the underlying data-generating process. This separation is similar to the approach in \citep{brown2021memorization}, which defines memorization using conditional mutual information between the dataset and the trained model, conditioned on the true concept. However, our notion differs in that it enables this separation at the instance level using algorithmic definitions of information.
 
To understand our new quantities, we measure unintended memorization and \intendedMemorizationName\ by training language models of varying capacity on datasets of different sizes. We first eliminate the question of generalization entirely by training on a dataset of random uniformly-sampled bitstrings. In this setting, we can exactly measure the amount of information contained about the data inside the model. This gives us a principled way to measure language model \textit{capacity} when trained on uniform datasets of exact known information content.  We find that GPT-style transformers can store between 3.5 and 4 bits of information in each model parameter, depending on model architecture and precision.

We then repeat our experiments with real text, where generalization is possible and even beneficial for learning. On real text, language models memorize up to a certain capacity, at which point they substitute unintended memorization for \intendedMemorizationName, and begin to learn general, reusable patterns as opposed to sample-level specifics. Our framework shows that double descent phenomenon begins to occur at this point, when the data size exceeds the model capacity in bits.

Finally, we use our results to predict a scaling law for membership inference performance based on model capacity and dataset size. We show that membership inference follows a clean relationship based on model capacity and dataset size: bigger models can memorize more samples, and making datasets bigger makes membership inference harder. Our scaling laws extrapolate to larger models, and  predict most modern language models are trained on too much data to do reliable membership inference on the average datapoint.

\begin{figure*}[t]
    \centering
    \begin{minipage}{0.47\textwidth}
        \centering
        \includegraphics[width=\textwidth]{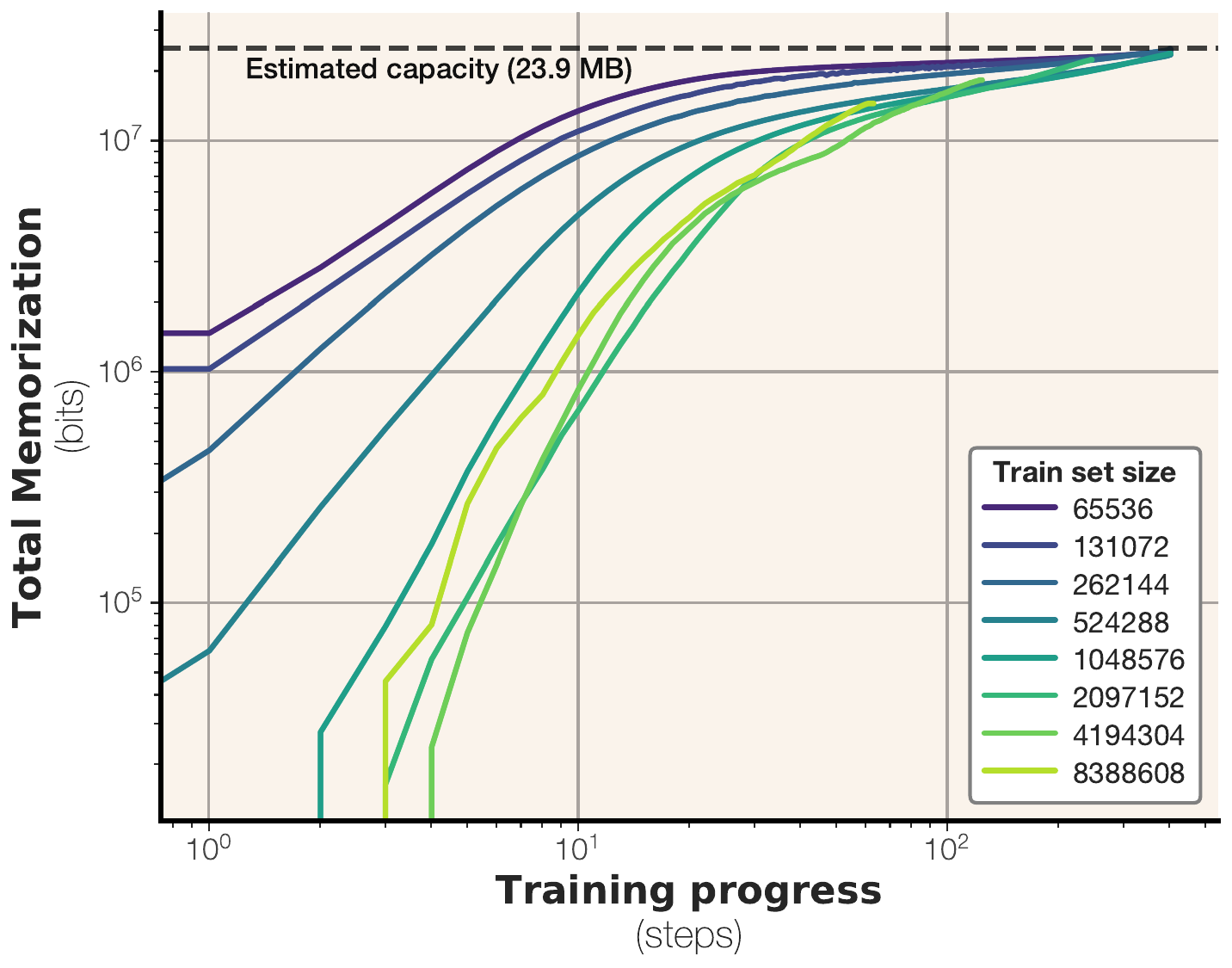}
        \caption{Bits memorized across training. This particular model is a GPT-style transformer with $6.86M$ parameters and a capacity of $23.9$ MB.}
        \label{fig:model_capacity_convergence}
    \end{minipage}
    \hfill
    \begin{minipage}{0.47\textwidth}
        \centering
        \includegraphics[width=\textwidth]{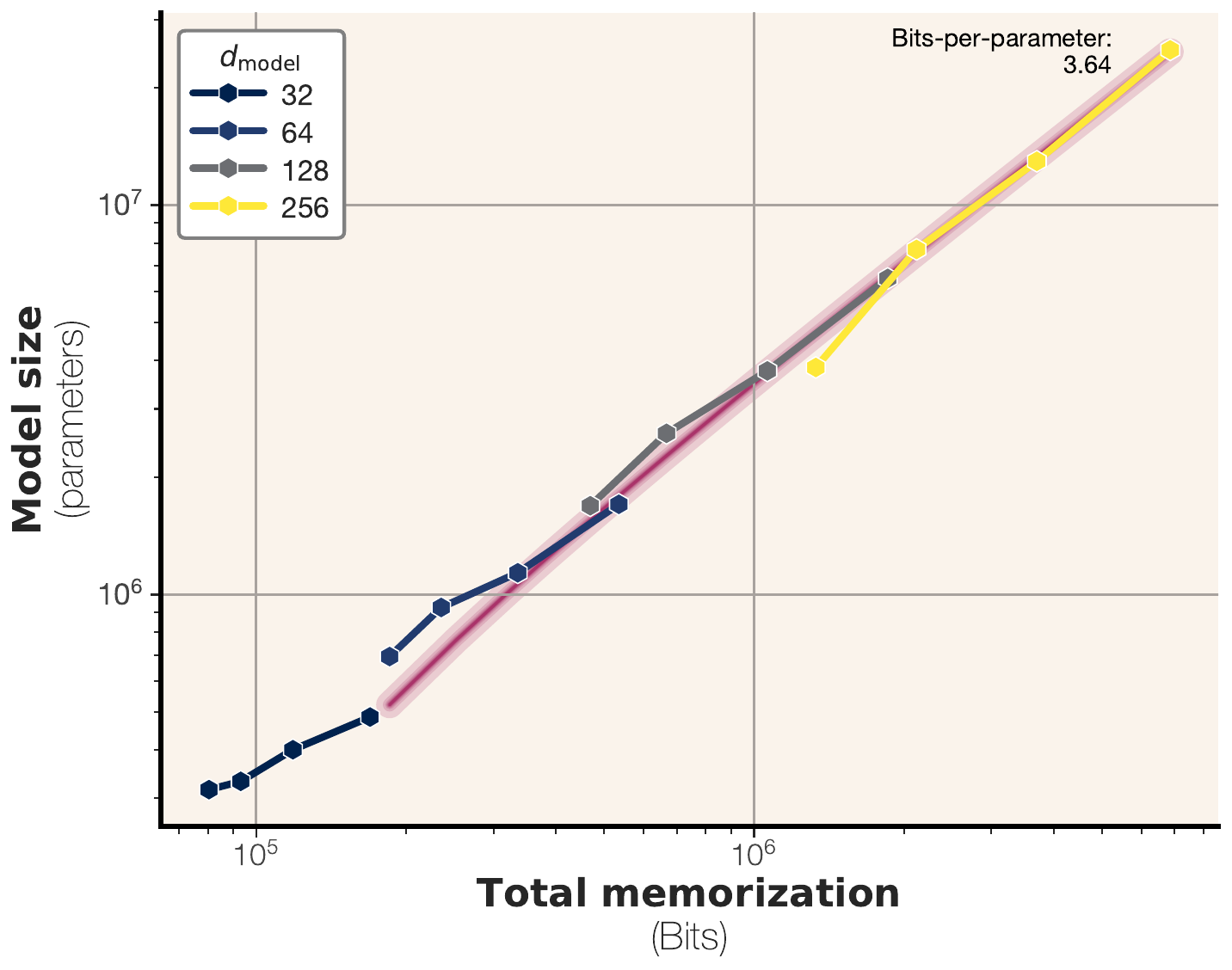}
        \caption{Capacity in bits-per-parameter for models trained on synthetic data. We estimate $\alpha = 3.64$ bits-per-parameter for GPT models trained in half precision.}
        \label{fig:model_capacity_parameters}
    \end{minipage}
\end{figure*}

\section{Memorization, intended and unintended}

When a model \(\theta = L(x)\) is trained using a training algorithm \(L\) and a dataset \(x \sim X\), some information is transferred from the sample \(x\) to the model \(\theta\). A key question in the memorization literature is determining how much of this stored information is intended versus unintended. 
In this work, we aim to provide a rigorous definition of memorization that satisfies certain properties:

\begin{enumerate}
\item{\textit{Separation from generalization.}} Our notion of unintended memorization must be distinct from intended memorization, which we refer to as generalization. For example, consider a language model trained on the sample:
   \emph{Q: What is \(2^{100}\)? A: 1267650600228229401496703205376.}
   When assessing how much of this training sample is memorized, we must account for the fact that performing simple math operations is expected from a language model. 
    \item{\textit{Sample-level memorization.}} We need to define memorization for realizations of random variables, not the random variables themselves. Specifically, we want to determine how much unintended memorization of a sample \(x\) occurs in a model \(\theta\). 
    
    \item{\textit{Independence from training algorithm.}} Our definition  should be independent of the training algorithm \(L\) and only a function of the final model \(\theta\) and the sample \(x\). This is crucial for language models, where we often only have access to the final model and target sample.
\end{enumerate}

Previous works have attempted to define memorization for machine learning models. 
We aim to provide precise definitions of memorization that meet our criteria, and offer ways to measure it. See Appendix \ref{app:memorization-related-work} for a broader discussion on definitions of memorization.

\subsection{Warm-up: A statistical view of memorization}\label{sec:shannon_memorization} 

\textit{Notation.} In this section, we use capital letters (e.g. $X$, $\Theta$) to refer to random variables and lowercase letters to refer to instances of a random variable (e.g. $x\sim X$ and $\theta\sim \Theta$).

We rely on information theory, which has developed well-understood notions of information for random variables. For a random variable $X$, we use $H(X)$, the entropy of $X$, to define the amount of information present in $X$. For two distinct random variables $X,Y$, we can define $X\mid Y$ as the uncertainty left in $X$ after fixing $Y$. With these definitions, we can now measure \textit{mutual information} between $X$ and $Y$ by subtracting the leftover information from the total information: $I(X,Y) = H(X) - H(X\mid Y)$.

Now suppose we have a machine learning pipeline. We have a prior  $\Theta$ on the underlying model that captures our dataset distribution $X$, and a learning algorithm $L$ that maps samples from $X$ to a trained model $\hat{\Theta}$.  To understand how much information about $X$ is stored in $\hat{\Theta}$, we can use the notion of mutual information:
\begin{align*}\text{mem} (X, \hat{\Theta})&= I(X, \hat{\Theta})= H(X) - H(X\mid \hat{\Theta}).
\end{align*}

Note that this captures all the information about $X$ that is stored in $\hat{\Theta}$. As we discussed, we need our notion of memorization to account for generalization as well. So when measuring unintended memorization, we are only interested in the information that is present in $X\mid \Theta$, which is the uncertainty left in $X$ after fixing $\Theta$.  Hence, we can define \textbf{unintended memorization} as
\begin{align*}
\text{mem}_U(X,\hat{\Theta},\Theta) = I([X\mid \Theta] , \hat{\Theta})
= H(X \mid \Theta) - H(X \mid (\Theta,\hat{\Theta})).
\end{align*} 

and then the \textbf{generalization} (or intended memorization) must be
\begin{align*}\text{mem}_I( X, \hat{\Theta},\Theta) = \text{mem}(X,\hat{\Theta}) - \text{mem}_U(X, \hat{\Theta}, \Theta)= I(X, \hat{\Theta}) - I ( [X \mid \Theta], \hat{\Theta})
\end{align*}

   having defined our notions of intended and unintended memorization we turn our attention to practically measuring them. Let us first state a proposition that enables measurement of unintended memorization:

\begin{proposition}[Super-additivity of Unintended Memorization]\label{prop:linearity_of_memorization}
    Assume $X=(X_1,\dots, X_n)$ is a dataset of $n$ i.i.d. samples. We have
    $$\sum_{i\in[n]} \text{mem}_U(X_i, \hat{\Theta}, \Theta)\leq \text{mem}_U(X, \hat{\Theta}, \Theta) \leq H(\hat{\Theta}).$$
\end{proposition}
This proposition shows that to measure a lower bound on the unintended memorization on the dataset level, we can sum per-sample memorization. On the other hand, the entropy of the information content of the trained model itself servers as an upper bound on the unintended memorization. Another implication of this implies that unintended memorization should scale with the dataset size but cannot exceed the total capacity of the model. 

We note that this statistical definition of memorization was first introduced by~\citet{brown2021memorization}, where they theoretically looked at the role of unintended memorization in enabling successful learning for certain tasks. However, this preliminary definition is unsuitable for us as we aim to practically measure instance-level memorization. In particular, since we observe only a single trained model and a single input sample, we are unable to define probabilities or conditional probabilities over samples conditioned on the model. 

\subsection{Measuring unintended memorization with Kolmogorov Complexity}

Our definitions of memorization and generalization so far are defined using an ``entropy-based" notion of information. This means our definitions can only be used for random variables. 
This brings big challenges in measuring memorization. All our variables in the definition of memorization are singletons. We have a single underlying model $\theta$, we have a single  dataset $x=(x_1,\dots, x_n)$ and we have a single trained model $\hat{\theta}$\footnote{Note the switch to lowercase variables because we are now working with instances, not random variables.}. It is impossible to measure the entropy (let alone conditional entropy) of the underlying variables using a single sample.

To this end, we switch to another notion of information based on compression, then later we show how this notion closely approximates the notion of memorization defined above. Kolmogorov complexity defines the information content of a string $x$, denoted as $H^K(x)$, to be the length of shortest representation of $x$ in a given computational model. Similarly, we can define the leftover information $x \mid \theta$, to be the shortest representation of $x$, when we have $\theta$ available as a reference. And the information content of $x \mid \theta$, denoted by $H^K(x \mid \theta)$, is the length of such description. Then, we can define mutual information in a similar fashion:
\begin{definition}[Kolmogorov complexity] Let $f$ be  an arbitrary computational model that takes a set of inputs and returns an output (e.g. universal Turing machine). The shortest description of  $x$ with respect to computational model $f$ is defined as
$H^K(x) = \min_{ f(p)=x} |p|.$ Also, the Kolmogorov complexity of $x$ relative to another string $\theta$ is defined as
$H^K(x\mid \theta) = \min_{ f(p, \theta)=x} |p|.$ 
And we define the Kolmogorov mutual information between $x$ and $\theta$ by
$I^K(x,\theta) = H^K(x) - H^K(x \mid \theta).$
 We assume inputs are bitstrings and $|p|$ is the bit length of the input.
\end{definition}
\begin{definition}[Kolmogorov memorization] 
Let $\theta$ be a reference model that approximates the true distribution of data, and $\hat{\theta}$ be a model trained on a dataset $x=(x_1,\dots, x_n)$. For each $x_i$ we define the memorization of $x_i$ in $\hat{\theta}$ as
$\text{mem}^K(\hat{\theta}, x) =  I^K(\hat{\theta}, x).$
We also define intended and unintended variants of memorization:
$$\text{mem}^K_U(x,\theta,\hat{\theta}) =H^K(x\mid \theta) - H^K(x\mid (\theta, \hat{\theta}))\text{,~and~mem}^K_I(x,\theta,\hat{\theta}) = \text{mem}^K(x,\hat{\theta}) - \text{mem}^K_U(x,\theta,\hat{\theta}).$$
\end{definition}

There are known connections between Kolmogorov complexity and Shannon Entropy \citep{grunwald2004shannoninformationkolmogorovcomplexity}. These results point at the conceptual connection between the two notions and imply that $\Ex_{x\sim X}[H^K(x)] \approx H(X)$. Interestingly, this implies that our notion of Kolmogorov memorization closely approximates Shannon memorization. 

\begin{proposition}[]\label{prop:kolmogorov_to_shannon}
    Let $X=(X_1,\dots,X_n)$ be an i.i.d, dataset distribution parametrized by ground-truth model $\theta$. Let $L$ be a training algorithm mapping $X$ to $\hat{\Theta}$. Assume $H(\hat{\Theta})= \ell$ and $H(X_i)=\ell'$  \footnote{The trained model and each data sample can be presented using $\ell$ and $\ell'$ bits respectively. }. 
    Then we have
    $\Big| \Ex_{\substack{x\sim X\\ \hat{\theta}\sim L(x)}}\big[\text{mem}^K_U(x_i,\hat{\theta}, \theta)] \big] -\text{mem}_{U}(X_i, \hat{\Theta}, \theta) \Big|  \leq \epsilon.$ for some constant $\epsilon$ independent of $ \ell, \ell'$ and $n$.
\end{proposition}

\subsection{Estimating Kolmogorov with compression}

Fixing our notion of Kolmogorov memorization, we now describe how we can estimate $H^K$ in different setups. Note that exact calculation of Kolmogorov complexity is known to be uncomputable \citep{kolmogorov1965three}. However, we can still approximate it using the best available compression schemes. These compression schemes could be arbitrary algorithms, e.g. using prompt optimization as in \cite{schwarzschild2024rethinkingllmmemorizationlens} or using text prefixes as in \cite{carlini2023quantifyingmemorizationneurallanguage}. 

We adopt arithmetic coding as the most natural compression algorithm for language. Arithmetic coding is not only effective for text compression \cite{deletang2024languagemodelingcompression}, but it also allows code lengths to be computed efficiently using model likelihoods. A promising direction for future research is to design compression algorithms specifically tailored to minimize the code length of training data in machine learning models and use it to obtain more accurate estimation of Kolmogorov  complexity and memorization. Below, we summarize how we approximate each term in our memorization definition using model likelihoods.

\begin{itemize}
    \item $H^K(x \mid \hat{\theta})$:
    Here, $\hat{\theta}$ is the trained target model, which does not necessarily capture the true data distribution. We do not really calculate the compressed code, instead we use fact that the compression rate of arithmetic coding tied to the likelihood of the model \citep{shannon1950englishentropy}. So, we can  estimate $H^K(x \mid \hat{\theta})$ by the negative log likelihood of $x$ under the target model, $-log(p(x \mid \hat{\theta}))$.
    \item $H^K(x \mid \hat{\theta}, \theta)$: In this case, the compression algorithm has access to both target and reference models. We simply compute $-log(\max\{p(x \mid \hat{\theta}), p(x \mid \theta)\})$. In practice, our choice of reference model is a larger model with the same architecture as $\theta$ trained for many steps on a much wider data distribution.
    
\end{itemize}
A curious reader may notice that we began with a likelihood-based notion of memorization, then shifted to a definition grounded in Kolmogorov complexity, and ultimately returned to likelihood to estimate that complexity. We emphasize, however, that the likelihood used in our approximation of Kolmogorov memorization is distinct from the initial likelihood notion. In particular, this likelihood is dependent on the parameters of the decoding algorithm, such as temperature or top-k sampling. More broadly, we note that any compression algorithm can be used to approximate Kolmogorov complexity, and our choice of arithmetic coding is just one instantiation of this broader framework.

\paragraph{Choice of reference model.} In this work we use two reference models to compute $p(x \mid \hat{\theta})$. In our experiments on synthetic random strings (\Cref{sec:um-synthetic}) we know the exact underlying data distribution and use that as a reference model. In our experiments on text (\Cref{sec:um-text}) we select $\theta$ to be a large model of the same family, trained on a much larger superset of the training data for $\hat{\theta}$.

\begin{table*}[t]
    \centering
    \begin{tabular}{@{}c@{\hspace{4pt}}c@{\hspace{4pt}}c|c@{\hspace{4pt}}c|c@{\hspace{4pt}}c@{}}
    \toprule
    & &  & \multicolumn{2}{c|}{$\text{Capacity}(\theta)$ \ \ \textit{\color{gray} [bits]}} & \multicolumn{2}{c}{$\alpha$ \ \ \textit{\color{gray} [bpp]}} \\
    $n_\text{layer}$ & $d_\text{model}$ & $\text{Params}$ &  fp32 & bf16 & fp32 & bf16 \\
    \midrule
    \multirow{4}{*}{1} 
        & 32 & 8.04$\times10^4$ & 3.39$\times10^5$ & 3.16$\times10^5$ & 4.23 & 3.93 \\
        & 64 & 1.85$\times10^5$ & 7.27$\times10^5$ & 6.93$\times10^5$ & 3.92 & 3.74 \\
        & 128 & 4.69$\times10^5$ & 1.71$\times10^6$ & 1.69$\times10^6$ & 3.65 & 3.61 \\
        & 256 & 1.33$\times10^6$ & 4.15$\times10^6$ & 3.83$\times10^6$ & 3.12 & 2.88 \\
        \midrule
    \multirow{4}{*}{2} 
        & 32 & 9.31$\times10^4$ & 3.87$\times10^5$ & 3.31$\times10^5$ & 4.16 & 3.56 \\
        & 64 & 2.35$\times10^5$ & 9.60$\times10^5$ & 9.27$\times10^5$ & 4.08 & 3.94 \\
        & 128 & 6.67$\times10^5$ & 2.66$\times10^6$ & 2.60$\times10^6$ & 3.99 & 3.89 \\
        & 256 & 2.12$\times10^6$ & 8.49$\times10^6$ & 7.76$\times10^6$ & 4.01 & 3.66 \\
       
        \midrule
    \multirow{4}{*}{4} 
        & 32 & 1.18$\times10^5$ & 4.65$\times10^5$ & 3.99$\times10^5$ & 3.92 & 3.37 \\
        & 64 & 3.35$\times10^5$ & 1.34$\times10^6$ & 1.14$\times10^6$ & 3.98 & 3.39 \\
        & 128 & 1.06$\times10^6$ & 4.02$\times10^6$ & 3.75$\times10^6$ & 3.78 & 3.53 \\
        & 256 & 3.70$\times10^6$ & 1.36$\times10^7$ & 1.30$\times10^7$ & 3.68 & 3.51 \\
        \midrule
    \multirow{4}{*}{8} 
        & 32 & 1.69$\times10^5$ & 5.12$\times10^5$ & 4.85$\times10^5$ & 3.02 & 2.86 \\
        & 64 & 5.35$\times10^5$ & 2.05$\times10^6$ & 1.71$\times10^6$ & 3.83 & 3.19 \\
        & 128 & 1.86$\times10^6$ & 7.23$\times10^6$ & 6.49$\times10^6$ & 3.89 & 3.49 \\
        & 256 & 6.86$\times10^6$ & 2.71$\times10^7$ & 2.51$\times10^7$ & 3.96 & 3.65 \\
    \midrule
        & & & & & \multicolumn{2}{c}{{Mean}\ ($\pm 0.1$):} \\
        & & & & & 3.83 & 3.51 \\
          
    \bottomrule
    \end{tabular}
    \caption{Model capacity estimates across different widths and depths in full and half-precision. Doubling precision from bfloat16 to float32 only increases model capacity from $3.51$ to $3.83$ bits-per-parameter.}
    \end{table*}
\hfill
\begin{figure}[htbp]
    \centering
    \begin{minipage}{0.47\textwidth}
        \centering
        \includegraphics[width=\textwidth]{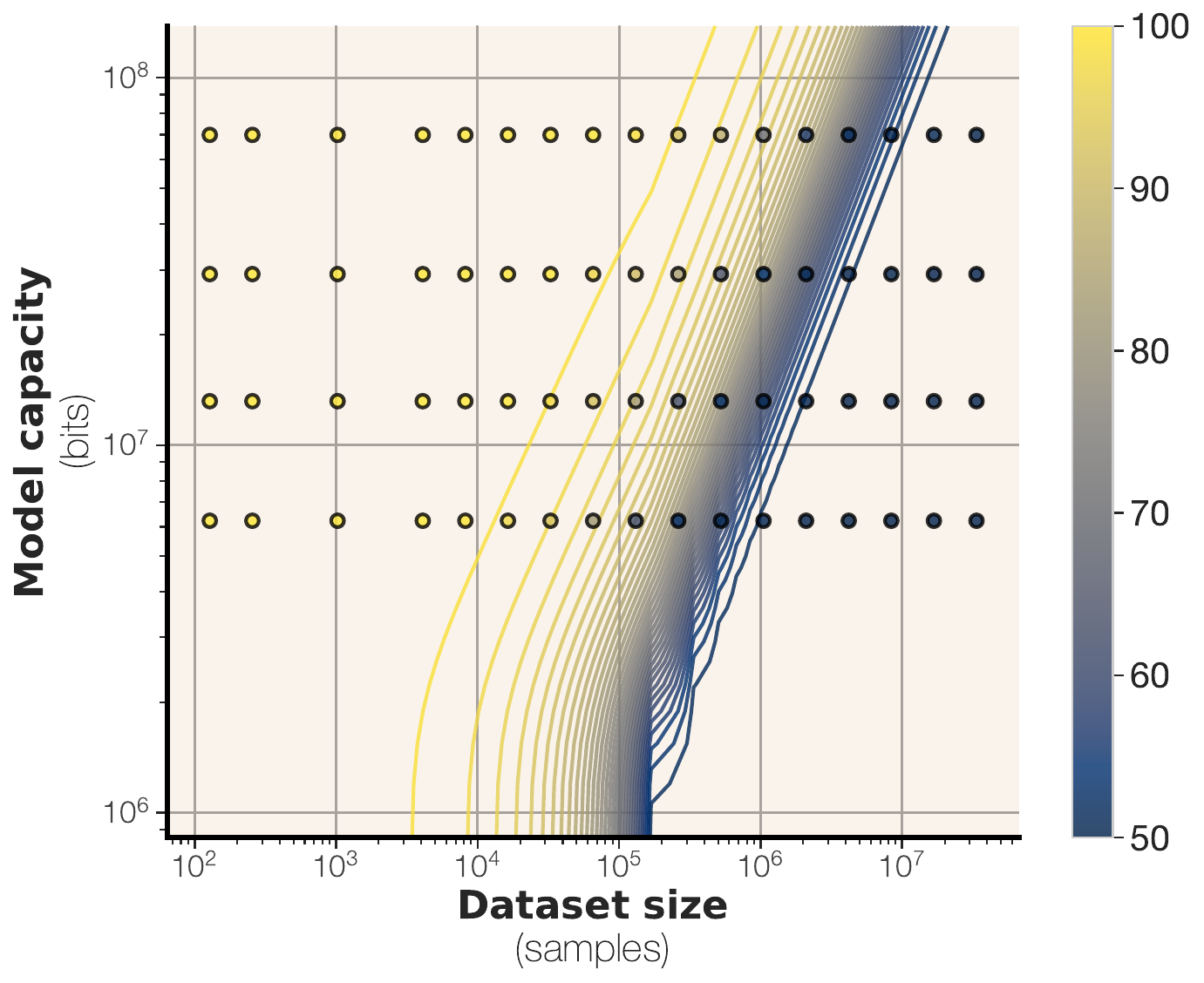}
        \caption{Scaling law curves for membership inference overlaid with empirical data, shown in circles.}
        \label{fig:memb-inf-heatmap}
    \end{minipage}
\end{figure}

\section{Model Capacity for Memorization}
\label{sec:um-synthetic}

\unintendedMemorizationNameCapital{} provides us a principled way of measuring the precise number of bits a model $\theta$ knows about a datapoint $x$. If we add up the information for each datapoint in a dataset, we can measure the total amount of bits a model knows about the dataset. And in cases where generalization is not possible because each datapoint is completely independent, we can estimate the \textbf{capacity} of a given model $\theta$ by summing per-datapoint \unintendedMemorizationName{}.

\subsection{Defining \textit{model capacity}}
We first formalize this notion of memorization capacity for a particular language model $\theta$. Capacity is the total amount of  memorization that can be stored in $\theta$ across all its parameters.
\begin{definition}[Capacity]
Let $X$ be a distribution and $L\colon  X\to \Theta$ a learning algorithm. We define the capacity of the learning algorithm $L$ to be 
\begin{align*}
\text{Capacity}(L) 
    &= \max_X \text{mem} \big(X,L(X)\big)
\end{align*}
\end{definition}

When the model capacity is reached, $\text{mem}(X, L(X))$ will no longer increase with dataset size. In practice, we can compute capacity by training to saturation on varying sizes of $X$ and computing the maximum memorization.

\subsection{Measuring model capacity with synthetic sequences}
\label{subsec:km-synthetic}
\label{subsec:um-synthetic}

In this section we measure the capacity of Transformer language models. Our goal is to instantiate multiple datasets and distributions and measure the memorization of them when training a single model $\theta$. Then, we take the maximum over all datasets to approximate of the model's capacity. For instantiating our datasets,  each token is uniformly sampled from a predefined set of tokens independent of the previous tokens. 

To approximate $H^k(x\mid \theta, \hat{\theta})$, we can directly compute entropy under the trained model to calculate the shortest description of the dataset conditioning on $\hat{\theta}$. Subtracting the two, we can approximate the \unintendedMemorizationName{}  $\text{mem}_U(X, L(X))$. Since the process for sampling the data is completely random, there is no \intendedMemorizationName{} to be stored within $\hat{\theta}$ (that is, $\text{mem}^U(X,L(X))\approx \text{mem}(X,L(X))$). 

Observe that when we sample synthetic sequences from a uniform distribution, we can compute their Shannon information exactly. Given a dataset size $N$, we construct a dataset of $N$ sequences, each of $S$ tokens. Given a vocabulary size $V$, we can calculate the total entropy of a dataset $x^i$ with such parameters by $H(x^i) = NS \log_2 V$. Then we calculate the compressed form $x^i$ using entropy under $\hat{\theta_i}$ to compute the code length and use this as an approximation of $H^K(x^i \mid \hat{\theta_j})$. Then we calculate the $\text{mem}(x^i, \hat{\theta_i})=H(x^i)-H^K(x^i \mid \hat{\theta_j})$ and compute a model's capacity as the maximum amount of memorization over all datasets.

\paragraph{Experimental details.}

In accordance with \citet{kaplan2020scalinglawsneurallanguage}, we train models with the GPT-2 architecture \citep{radford2019language} initialized from scratch. Our models have between $1$ and $8$ layers, hidden dimensions scaled from $32$ to $512$, and from $100K$ to $20M$ parameters. We train models for $10^6$ steps with a batch size of $2048$. We use the Adam optimizer. All models are trained on a single A100 GPU in bfloat16 precision, and we use gradient accumulation if a batch cannot fit in memory. Unless otherwise noted, we set vocabulary size $V = 2048$, sequence length $S = 64$ and vary only the number of points in a dataset. We train each model on each dataset size over five random seeds, which affect both model initialization and the dataset sampling.

\paragraph{Results.}
We plot memorization across model and data sizes in Figure \ref{fig:model_capacity_synthetic}. This allows us to visualize \unintendedMemorizationName{} amounts (y-axis) across dataset sizes (x-axis) grouped by model size (line color). We observe a striking plateau once a model reaches its capacity. Given the dataset is large enough, models exhibit an upper bound in net memorization, regardless of data size. Small datasets are completely memorized by all models with enough capacity.

We estimate the capacity of each model as the maximum amount of \unintendedMemorizationName{} in bits measured across all dataset sizes. We then compare this capacity to the model size in Figure \ref{fig:model_capacity_parameters}. Interestingly, even at this small scale, we see a very smooth relationship between observed capacity (maximum memorization measured over all datasets) and model parameters. We plot this relationship in Figure \ref{fig:model_capacity_parameters}: under these settings, our models consistently memorize between $3.5$ and $3.6$ bits per parameter. This corroborates the findings of prior work such as \citep{roberts2020knowledgepackparameterslanguage,lu2024scalinglawsfactmemorization}, which noticed that fact storage scales linearly with model capacity. Ours is a slightly larger estimate than \citet{allenzhu2024physicslanguagemodels33}, which estimated via quantization that models can store around $2$ bits per parameter.

Since our models are learned via gradient descent, they are not guaranteed to find the global optima; thus, we are only ever measuring a lower bound on model capacity. We take a closer look at the training curves to analyze the convergence of our 8M parameter language model. We plot model convergence throughout training in Figure \ref{fig:model_capacity_convergence}.

In this case, all datasets from 16,000 to 4M samples fall within a range of $3.56-3.65 \times 10^6$ bits memorized. This indicates that our measurements are robust within an order of magnitude, and we do not expect to memorize significantly more information by training for more steps. This finding also confirms our hypothesis that capacity scales roughly with parameter count. The two largest datasets (4M and 8M samples, respectively) converge to total memorization of $2.95\times 10^6$ and $1.98\times 10^6$  bits memorized. We expect that their memorization rates would continue to increase toward the capacity had we trained for more epochs.

\paragraph{How does precision affect capacity?} One natural question is how our estimates for $\alpha$ depend on the precision of language model training. In fact, although most software defaults to training in 32-bit precision, recent work has shown that language models can be quantized to fewer than 2 bits per parameter and still retain much of their utility. Since all other experiments have been conducted in bfloat16 precision, rerun our experiments in full fp32 precision to analyze the effect on capacity. Across model sizes, we observe a small increase in capacity, and an increase in $\alpha$ from 3.51 to $3.83$ bits-per-parameter on average. This is far less than the actual 2x increase in the bits of $\theta$, indicating that \textbf{most of the extra model bits added when increasing precision from bfloat16 to float32 are not used} for raw storage.

\section{Disentangling Unintended Memorization from Generalization}
\label{sec:um-text}

Our previous experiments analyzed the memorization and membership inference properties of synthetic bitstrings. We now turn to measuring memorization of text. Unlike randomly generated sequences, learning from text data is a mix of both \unintendedMemorizationName{} (sample-level) and \intendedMemorizationName{} (population-level). 
Therefore, as a reference model, we use the model of an equal parameter count trained on the maximum amount of data (in this case, the entire dataset).\footnote{Restricting the computational power of the reference model relates its predictions to $\mathcal{V}$-information \citep{xu2020vinformation} which measures the ``usable'' information available in a signal, when accounting for model size.} We also consider an \textit{oracle} reference model, which is the model that achieves the best compression rate (lowest loss) on the evaluation dataset, and may have many more parameters.

\paragraph{Experimental details.} We repeat the experiments from \Cref{subsec:km-synthetic}, substituting our synthetic datapoints for real text. To obtain a distribution of real-world text data, we could use any pre-training scale text dataset; we use the recently proposed FineWeb dataset \citep{penedo2024fineweb} as it follows state-of-the-art deduplication practices. We use sequences of $64$ tokens but perform an additional deduplication step to ensure perfect deduplication (otherwise, that $1-2\%$ of sequences become duplicates when truncating to $64$ tokens). We find careful deduplication extremely important for faithfully measuring extraction rates. As in the previous subsection, we pretrain models of varying sizes on different-sized text datasets and measure the \unintendedMemorizationName{} of each model-dataset pair. In addition to memorization, we measure membership inference performance according to a standard loss-based membership inference procedure; we also compute exact extraction rates by greedily decoding prefixes of different lengths.

\paragraph{Results.} We first observe that the sample-level \unintendedMemorizationName{} increases with model parameters and decreases with training set size (Figure \ref{fig:text-loss-train-val}). When we measure \unintendedMemorizationName{} with respect to an oracle reference model (Figure \ref{fig:text-kolmogorov-total-oracle}), memorization steadily increases as our smaller model is able to learn more about the small training set than the oracle, and then decreases as our model starts to generalize and perform on average worse than the (higher-capacity) oracle.

\paragraph{Dataset-to-capacity ratio predicts double descent.} We observe from the train and test loss that for larger datasets the model only begins to generalize (i.e. evaluation loss decreases) once its capacity is reached, which takes approximately $10^5$ samples, depending on parameter count. As in \citet{nakkiran2019deepdoubledescentbigger} we plot the ratio between the dataset size and model capacity (Figure \ref{fig:synth_double_descent}). Unlike prior work, in our experiments we can compute the exact dataset size (based on the compression rates of the reference model) and exact model capacity (based on our estimate of $\alpha$). 

We clearly observe double descent evaluation performance decreases as the training set size nears model capacity, and then rapidly drops as the dataset capacity exceeds the capacity of the model.  Our observations offer an intuitive explanation for double descent \citep{belkin2019biasvariance,nakkiran2019deepdoubledescentbigger}: \textbf{double descent begins exactly when the data capacity exceeds the model capacity}. One theory is that once the model can no longer memorize datapoints individually, it is forced to share information between datapoints to save capacity, which leads to generalization.

\begin{figure*}
    \begin{minipage}[t]{0.32\textwidth}
        \centering
        \includegraphics[width=\textwidth]{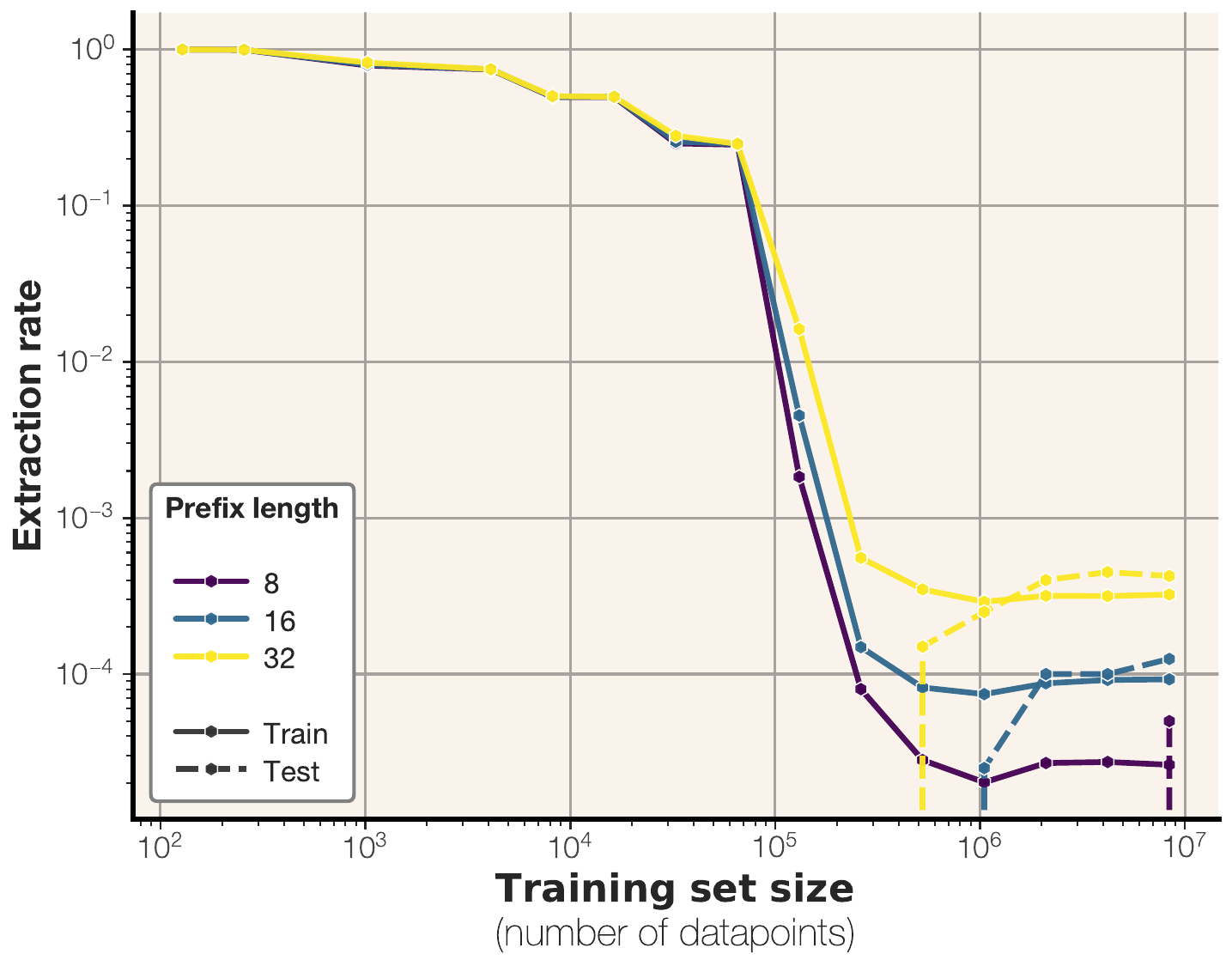}
        \caption{Extraction rates of 64-token training sequences across prefix lengths, for both train and evaluation.}
        \label{fig:text-extraction-rates}
    \end{minipage}
    \hfill
    \begin{minipage}[t]{0.32\textwidth}
        \centering
        \includegraphics[width=\textwidth]{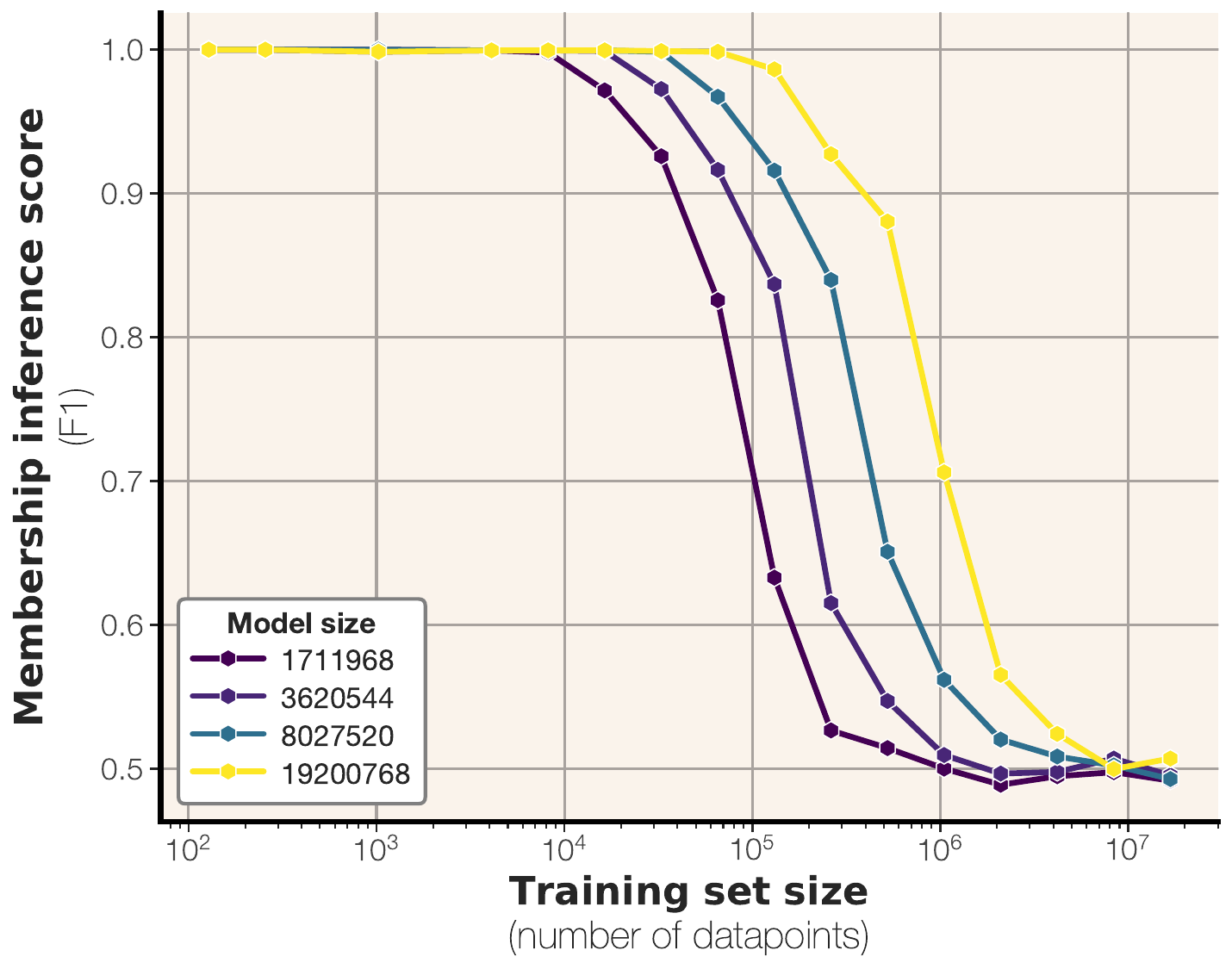}
        \caption{Membership inference F1 across dataset sizes. In this case, F1 score of 0.5 implies random guessing.      
}

        \label{fig:text-membership}
    \end{minipage}
    \hfill
    \begin{minipage}[t]{0.32\textwidth}
        \centering
        \includegraphics[width=\textwidth]{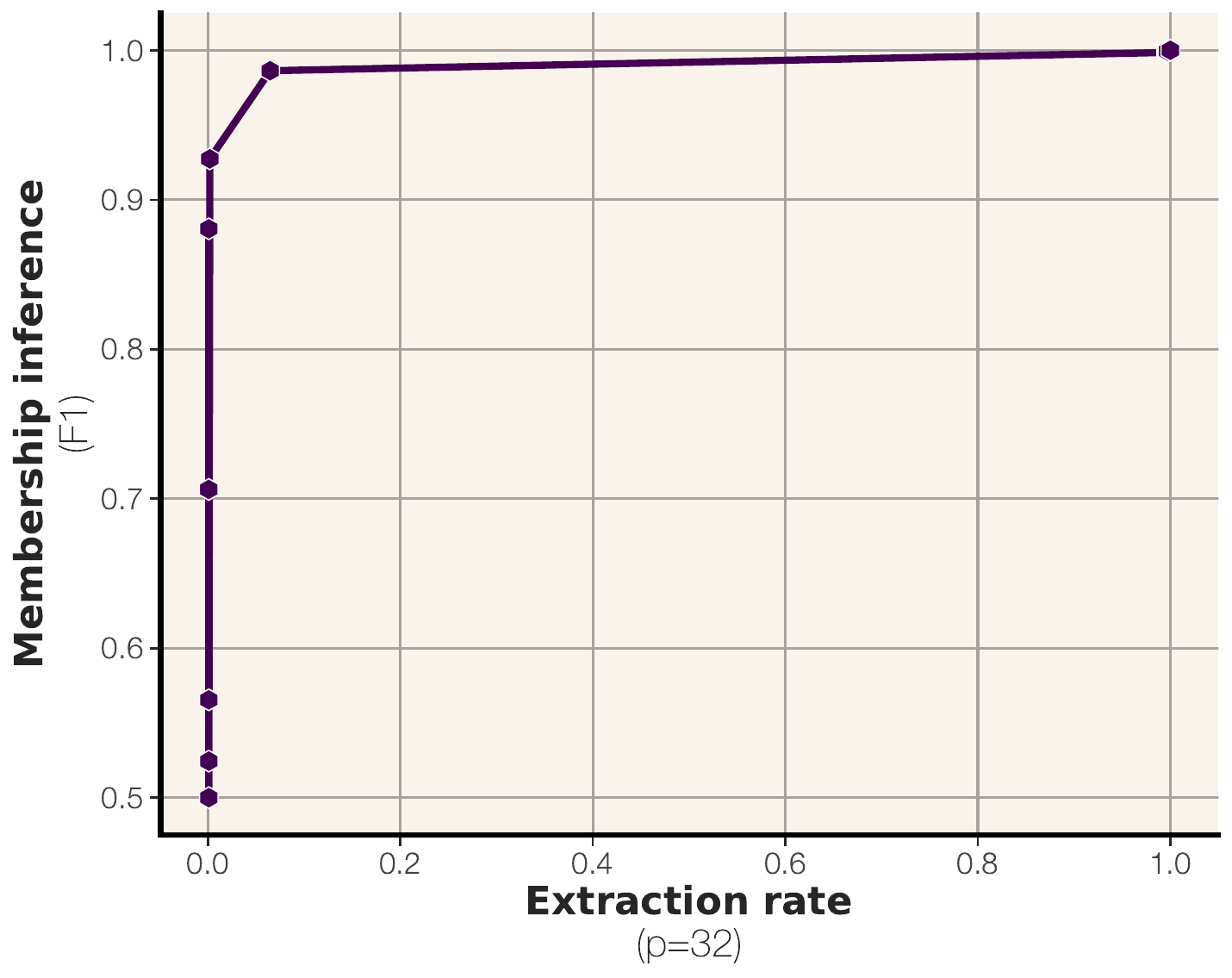}
        \caption{Membership inference vs 32-token-prefix suffix extraction rate. Membership inference is generally easier than extraction.}
        \label{fig:text-membership-vs-extraction}
    \end{minipage}
    \vspace{-10pt}
\end{figure*}

\paragraph{Generalization explains nonzero extraction rates.} We measure extraction rates on the full training set and 10,000 non-overlapping test samples (Figure \ref{fig:text-extract-eval}). We note that for 32-token prefixes, 100\% are extractable for very small training set sizes; predictably, all extraction numbers decrease with training set size. When the dataset sizes grows sufficiently large, the extraction rate does not go fully to zero; however, it converges to nearly exactly the test extraction rate. In other words, when our (deduplicated) dataset grows sufficiently large, \textbf{all successful training data extraction is attributable to generalization}.

\section{Memorization and Membership}

Our training settings allow total control over the train and test data and come with perfect deduplication. This makes our setting ideal for studying the relationship between model size, dataset size, and membership inference success rate.

All of our membership inference results come from a standard loss-based membership inference \citep{yeom2018privacyriskmachinelearning, sablayrolles2019whiteboxvsblackboxbayes}. The method is very simple: we set a cutoff loss value to predict whether a sample is or is not a member of the training dataset.

\subsection{Membership in synthetic and text data}

\paragraph{Synthetic data.} For each of our models trained on synthetic data, we plot the success rate of the membership inference attack attack across dataset sizes. We show results in Figure \ref{fig:model_capacity_membership}. Above a certain dataset size, membership inference starts to fail in the average case.  This finding indicates that if the dataset size is too large compared to the model, membership inference of an average training sample may not be possible.

\paragraph{Text.} For each of our models trained on text, we use unused non-overlapping data from FineWeb to perform a standard loss-based membership inference \citep{yeom2018privacyriskmachinelearning, sablayrolles2019whiteboxvsblackboxbayes} on each model and plot performance across dataset sizes (\ref{fig:text-membership}). For a fixed model size, membership inference gets more difficult as the size of the data increases. When comparing membership inference to extraction (Figure \ref{fig:text-membership-vs-extraction}), membership inference is strictly higher in every case; in some cases we can infer training dataset membership quite well (score of $0.97$) with an extraction rate of $0$.

\subsection{Scaling laws for Membership}
\label{sec:scaling-laws-membership}
In this section we develop a set of predictive models for memorization. Specifically, we predict the F1 score of a loss-based membership attack given token count, number of examples, and model parameter count. We then validate our predictions on models from $500$K to $1.5$B parameters.

\subsubsection{Functional forms}

We observe that for a fixed model capacity, membership inference follows a roughly sigmoidal form with respect to dataset size. The intuitive explanation is that M.I. is easy for large models overfit to tiny datasets, so its score begin at 1; as dataset size increases, differentiating train from test data by loss becomes more and more difficult, eventually decaying toward 0.5.

We reuse the data collected in our text experiments (Section \ref{sec:um-text}) to solve for constants $c_1, c_2, c_3$ in the following equation:
\begin{equation*}
    \text{Membership}_{F_1}(\theta, \mathcal{D}) = \dfrac{1}{2}{(1 + c_1\sigma(c_2 (\dfrac{\text{Capacity}(\theta)}{|\mathcal{D}|} + c_3))}
\end{equation*}
where $\sigma(x) = \frac{1}{1 + e^{-x}}$.

\paragraph{Limiting behavior.} We observe that as $|\mathcal{D}| \rightarrow \infty$, performance of our membership inference attack decreases to $0.5$ (essentially random performance). For a model trained on an infinite dataset, our law predicts both membership inference and extraction to be impossible.
\vspace{-10pt}
\paragraph{Fitting.} We use a non-linear least squares solver to find optimal values for $c_1, c_2, c_3$. Solutions found are $c_1 = 1.34$, $c_2=-0.034$, and $-33.14$. We plot the scaling laws along with observed data in Figure \ref{fig:memb-inf-heatmap}. Although the sigmoidal function is slightly simplistic (the points do not perfectly fit) our fit produces estimates within $1-2\%$ of observations.

\subsubsection{Validation on larger models}
We note that all contemporary language models trained with a tokens-per-parameter ratio of $10^2$ or higher, which according to our laws would imply membership inference score of $0.5$ – that is, within our formulation, statistically significant loss-based membership inference is not possible.

\begin{table*}[]
    \centering
    \begin{tabular}{c|ccc|c|c|c}
    \toprule
     & $d_{emb}$   & $n_{layer}$  & $|\theta|$ & $|D|$ & Predicted F1 & Observed F1 \\ \midrule
    \multirow{3}{*}{GPT2-XL} & \multirow{3}{*}{1600} & \multirow{3}{*}{48} & \multirow{3}{*}{1,556,075,200} & 170,654,583  & 0.55 &  $54.61 \pm 1.3$  \\
    &  &   &  & 76,795,021 & 0.75 &  $71.08 \pm 0.4$ \\
    & &   &  & 18,851,574 & 0.95 & $95.85 \pm 0.8$ \\
     \midrule
    \multirow{3}{*}{GPT2-Medium} & \multirow{3}{*}{768}  & \multirow{3}{*}{12} & \multirow{3}{*}{123,702,528}   & 13,566,442  & 0.55  & $53.44 \pm 1.1$  \\
     & &   &  & 6,104,935 & 0.75 & $65.69 \pm 0.6$ \\
     & &   &  & 1,498,634 & 0.95 &  $97.98 \pm 0.3$ \\
     \bottomrule
    \end{tabular}
    \caption{Dataset sizes that our scaling law predicts will produce a given membership inference F1, along with empirical values.}
    \label{tab:scaling-law-dataset-sizes}
\end{table*}

To validate our predictions, we train models with expected membership $F1$ scores of 0.55, 0.75, and 0.95. For model sizes we select GPT-2 small ($125$M params) and GPT-2 XL ($1.5$B params). Using our scaling law, we solve for the dataset size required to get the desired membership inference score for the given model size (see Table \ref{tab:scaling-law-dataset-sizes} for more information). We train models on the estimated dataset size and measure F1 score (shown as circles in Figure \ref{fig:memb-inf-heatmap}). 

Our predictions are generally within $1.5$ points of the true F1 score; the score is most inaccurate for estimated F1 of 0.75, which is the point where the sigmoid is steepest. In general, the accuracy of our results indicates that our empirical model of membership inference is relatively accurate and provides evidence for why membership inference attacks fail on models trained on extremely large datasets \citep{das2024blindbaselinesbeatmembership, duan2024membership, maini2024llmdatasetinferencedid}. 

\section{Related Work}

\paragraph{Language models and compression.} Shannon's source coding theorem \citep{shannon1948noisychannel} first formalized the duality between prediction and compression. The connection between language modeling and compression was studied as far back as \citet{shannon1950englishentropy}, which observed that more accurate models of English can compress text in fewer bits. Other works note the connection between Kolmogorov complexity \citep{kolmogorov1965three} and Shannon information in detail \citep{grunwald2004shannoninformationkolmogorovcomplexity}.
\citet{deletang2024languagemodelingcompression} investigate using modern transformer-based language models as compressors. We use compression as a tool to measure memorization in models.

\paragraph{Language model capacity.}  Early research on single-layer perceptrons found that single-layer networks can store up to 2 bits-per-parameter \citep{cover1965geometrical, gardner1988space, baldi1989neural}. 
\citep{arpit2017closerlookmemorizationdeep} formalize the idea of \textit{effective capacity} of a model and its training procedure; they also observe that both representation capacity and training time have a strong impact on empirical model capacity. Several other works measure language model capacity in the number of facts or random labels that can be memorized by a network such as an RNN \citep{collins2017capacityrnns, boo2019memcapacity} or transformer \citep{roberts2020knowledgepackparameterslanguage, heinzerling2021languagemodelsknowledgebases, allenzhu2024physicslanguagemodels33}, sometimes under quantization. A few research efforts
\citep{yun2019smallrelunetworkspowerful, curth2023uturndoubledescentrethinking, mahdavi2024memorizationcapacitymultiheadattention, kajitsuka2024optimalmemorizationcapacitytransformers} have developed theoretical estimates for the capacity of different model architectures, although none have yet scaled to multi-layer modern transformers. \citep{shwartzziv2024justflexibleneuralnetworks} also analyze the `capacity' of neural networks on datasets of varying size.
We are the first to measure a clear upper-bound in model capacity using per-sample entropy measurements.


\paragraph{Information regularization in learning theory.} Several works have explored the role of mutual information between the input and output of a learning algorithm~\citep{bassily2018learners, haghifam2020sharpened,steinke2020reasoning}. This concept closely relates to the notion of memorization based on Shannon information, discussed in Section~\ref{sec:shannon_memorization}. Some of our findings also relate to the discovery of \textit{double descent} in machine learning  \citep{belkin2019biasvariance, nakkiran2019deepdoubledescentbigger} and language modeling \citep{xia2023trainingtrajectorieslanguagemodels}, as well as general discussions of memorization and generalization in deep learning \citep{zhang2017understandingdeeplearningrequires, tanzer2022memorisationversusgeneralisationpretrained}.

\paragraph{Alternative definitions of memorization.} Unintended memorization is deeply related to the many other definitions of memorization proposed in the literature. We provide a detailed comparison in the following subsections.


\paragraph{Prior definitions of memorization.} \citet{carlini2019secretsharerevaluatingtesting} defined a string \( m \) as memorized by a language model \( \theta \) if the second half of \( m \) can be generated greedily when prompting the model with the first half. Following this, \citet{nasr2023scalableextractiontrainingdata} introduced \textit{extractable memorization}, where model \( \theta \) is said to memorize \( m \) if an adversarial prompt \( p \) can be found that generates \( m \). \citet{mireshghallah2022memorizationnlpfinetuningmethods} and \citet{schwarzschild2024rethinkingllmmemorizationlens} refined this definition by restricting \( p \) to a certain number of tokens, preventing it from containing the entire \( m \). However, even this definition has limitations: for example, generating the sequence ``cat cat cat ... cat" with the prompt "repeat cat 1000 times" does not necessarily indicate memorization. \citet{carlini2019secretsharerevaluatingtesting} use perplexity or likelihood, one measure of the compressibility of a sequence, in an effort to distinguish highly memorized sequences from merely easy-to-compress ones. One additional definition of note is \textit{counterfactual memorization} \citep{zhang2023counterfactualmemorizationneurallanguage}, which measures the impact of a single datapoint on training; this can be seen as an instantiation of our definition where a different model of the same family is used as a reference model. Overall, all these works regarded memorization in terms that can be seen as forms of compression, although did not explicitly define it as such.

Finally, a concurrent work \citep{cohen2024datareconstructiondont} proposes a theoretical definition for memorization also relying on Kolmogorov.

\section{Conclusion}
We propose a new definition of memorization that allows us to measure the exact number of bits a model knows about a dataset. We use our definition to measure the capacity of modern transformer language models and analyze how measurements such as extraction and F1 score scale with model and dataset size. We also propose a scaling law for membership inference and validate it on larger models. Our results help further practitioner understanding of how language models memorize and what they might (or might not) be memorizing across model and dataset scales.

\section{Acknowledgements}

Thanks to the many folks who helped us improve our paper, including Karen Ullrich, Niloofar Mireshghallah, Mark Ibrahim, Preetum Nakkiran, and Léon Bottou. 


\bibliography{refs}

\begin{thebibliography}{57}
\providecommand{\natexlab}[1]{#1}
\providecommand{\url}[1]{\texttt{#1}}
\expandafter\ifx\csname urlstyle\endcsname\relax
  \providecommand{\doi}[1]{doi: #1}\else
  \providecommand{\doi}{doi: \begingroup \urlstyle{rm}\Url}\fi

\bibitem[Allen-Zhu \& Li(2024)Allen-Zhu and Li]{allenzhu2024physicslanguagemodels33}
Allen-Zhu, Z. and Li, Y.
\newblock Physics of language models: Part 3.3, knowledge capacity scaling laws, 2024.
\newblock URL \url{https://arxiv.org/abs/2404.05405}.

\bibitem[Arpit et~al.(2017)Arpit, Jastrzebski, Ballas, Krueger, Bengio, Kanwal, Maharaj, Fischer, Courville, Bengio, and Lacoste-Julien]{arpit2017closerlookmemorizationdeep}
Arpit, D., Jastrzebski, S., Ballas, N., Krueger, D., Bengio, E., Kanwal, M.~S., Maharaj, T., Fischer, A., Courville, A., Bengio, Y., and Lacoste-Julien, S.
\newblock A closer look at memorization in deep networks, 2017.
\newblock URL \url{https://arxiv.org/abs/1706.05394}.

\bibitem[Baldi \& Hornik(1989)Baldi and Hornik]{baldi1989neural}
Baldi, P. and Hornik, K.
\newblock Neural networks and principal component analysis: Learning from examples without local minima.
\newblock \emph{Neural Networks}, 2\penalty0 (1):\penalty0 53--58, 1989.

\bibitem[Bassily et~al.(2018)Bassily, Moran, Nachum, Shafer, and Yehudayoff]{bassily2018learners}
Bassily, R., Moran, S., Nachum, I., Shafer, J., and Yehudayoff, A.
\newblock Learners that use little information.
\newblock In \emph{Algorithmic Learning Theory}, pp.\  25--55. PMLR, 2018.

\bibitem[Belkin et~al.(2019)Belkin, Hsu, Ma, and Mandal]{belkin2019biasvariance}
Belkin, M., Hsu, D., Ma, S., and Mandal, S.
\newblock Reconciling modern machine-learning practice and the classical bias–variance trade-off.
\newblock \emph{Proceedings of the National Academy of Sciences}, 116\penalty0 (32):\penalty0 15849–15854, July 2019.
\newblock ISSN 1091-6490.
\newblock \doi{10.1073/pnas.1903070116}.
\newblock URL \url{http://dx.doi.org/10.1073/pnas.1903070116}.

\bibitem[Bhattacharjee et~al.(2023)Bhattacharjee, Dasgupta, and Chaudhuri]{bhattacharjee2023data}
Bhattacharjee, R., Dasgupta, S., and Chaudhuri, K.
\newblock Data-copying in generative models: a formal framework.
\newblock In \emph{International Conference on Machine Learning}, pp.\  2364--2396. PMLR, 2023.

\bibitem[Boo et~al.(2019)Boo, Shin, and Sung]{boo2019memcapacity}
Boo, Y., Shin, S., and Sung, W.
\newblock Memorization capacity of deep neural networks under parameter quantization, 05 2019.

\bibitem[Brown et~al.(2021)Brown, Bun, Feldman, Smith, and Talwar]{brown2021memorization}
Brown, G., Bun, M., Feldman, V., Smith, A., and Talwar, K.
\newblock When is memorization of irrelevant training data necessary for high-accuracy learning?
\newblock In \emph{Proceedings of the 53rd annual ACM SIGACT symposium on theory of computing}, pp.\  123--132, 2021.

\bibitem[Carlini et~al.(2019)Carlini, Liu, Úlfar Erlingsson, Kos, and Song]{carlini2019secretsharerevaluatingtesting}
Carlini, N., Liu, C., Úlfar Erlingsson, Kos, J., and Song, D.
\newblock The secret sharer: Evaluating and testing unintended memorization in neural networks, 2019.
\newblock URL \url{https://arxiv.org/abs/1802.08232}.

\bibitem[Carlini et~al.(2023{\natexlab{a}})Carlini, Hayes, Nasr, Jagielski, Sehwag, Tramèr, Balle, Ippolito, and Wallace]{carlini2023extractingtrainingdatadiffusion}
Carlini, N., Hayes, J., Nasr, M., Jagielski, M., Sehwag, V., Tramèr, F., Balle, B., Ippolito, D., and Wallace, E.
\newblock Extracting training data from diffusion models, 2023{\natexlab{a}}.
\newblock URL \url{https://arxiv.org/abs/2301.13188}.

\bibitem[Carlini et~al.(2023{\natexlab{b}})Carlini, Ippolito, Jagielski, Lee, Tramer, and Zhang]{carlini2023quantifyingmemorizationneurallanguage}
Carlini, N., Ippolito, D., Jagielski, M., Lee, K., Tramer, F., and Zhang, C.
\newblock Quantifying memorization across neural language models, 2023{\natexlab{b}}.
\newblock URL \url{https://arxiv.org/abs/2202.07646}.

\bibitem[Cohen et~al.(2024)Cohen, Kaplan, Mansour, Moran, Nissim, Stemmer, and Tsfadia]{cohen2024datareconstructiondont}
Cohen, E., Kaplan, H., Mansour, Y., Moran, S., Nissim, K., Stemmer, U., and Tsfadia, E.
\newblock Data reconstruction: When you see it and when you don't, 2024.
\newblock URL \url{https://arxiv.org/abs/2405.15753}.

\bibitem[Collins et~al.(2017)Collins, Sohl-Dickstein, and Sussillo]{collins2017capacityrnns}
Collins, J., Sohl-Dickstein, J., and Sussillo, D.
\newblock Capacity and trainability in recurrent neural networks, 2017.
\newblock URL \url{https://arxiv.org/abs/1611.09913}.

\bibitem[Cover(1965)]{cover1965geometrical}
Cover, T.~M.
\newblock Geometrical and statistical properties of systems of linear inequalities with applications in pattern recognition.
\newblock \emph{IEEE Transactions on Electronic Computers}, EC-14\penalty0 (3):\penalty0 326--334, 1965.

\bibitem[Curth et~al.(2023)Curth, Jeffares, and van~der Schaar]{curth2023uturndoubledescentrethinking}
Curth, A., Jeffares, A., and van~der Schaar, M.
\newblock A u-turn on double descent: Rethinking parameter counting in statistical learning, 2023.
\newblock URL \url{https://arxiv.org/abs/2310.18988}.

\bibitem[Das et~al.(2024)Das, Zhang, and Tramèr]{das2024blindbaselinesbeatmembership}
Das, D., Zhang, J., and Tramèr, F.
\newblock Blind baselines beat membership inference attacks for foundation models, 2024.
\newblock URL \url{https://arxiv.org/abs/2406.16201}.

\bibitem[Delétang et~al.(2024)Delétang, Ruoss, Duquenne, Catt, Genewein, Mattern, Grau-Moya, Wenliang, Aitchison, Orseau, Hutter, and Veness]{deletang2024languagemodelingcompression}
Delétang, G., Ruoss, A., Duquenne, P.-A., Catt, E., Genewein, T., Mattern, C., Grau-Moya, J., Wenliang, L.~K., Aitchison, M., Orseau, L., Hutter, M., and Veness, J.
\newblock Language modeling is compression, 2024.
\newblock URL \url{https://arxiv.org/abs/2309.10668}.

\bibitem[Duan et~al.(2024)Duan, Suri, Mireshghallah, Min, Shi, Zettlemoyer, Tsvetkov, Choi, Evans, and Hajishirzi]{duan2024membership}
Duan, M., Suri, A., Mireshghallah, N., Min, S., Shi, W., Zettlemoyer, L., Tsvetkov, Y., Choi, Y., Evans, D., and Hajishirzi, H.
\newblock Do membership inference attacks work on large language models?
\newblock In \emph{Conference on Language Modeling (COLM)}, 2024.

\bibitem[Dubey \& et~al(2024)Dubey and et~al]{dubey2024llama3herdmodels}
Dubey, A. and et~al, A.~J.
\newblock The llama 3 herd of models, 2024.
\newblock URL \url{https://arxiv.org/abs/2407.21783}.

\bibitem[Dwork(2006)]{dwork2006differential}
Dwork, C.
\newblock Differential privacy.
\newblock In \emph{International Colloquium on Automata, Languages, and Programming}, pp.\  1--12. Springer, 2006.

\bibitem[Feldman(2020)]{feldman2020does}
Feldman, V.
\newblock Does learning require memorization? a short tale about a long tail.
\newblock In \emph{Proceedings of the 52nd Annual ACM SIGACT Symposium on Theory of Computing}, pp.\  954--959, 2020.

\bibitem[Gardner(1988)]{gardner1988space}
Gardner, E.
\newblock The space of interactions in neural network models.
\newblock \emph{Journal of Physics A: Mathematical and General}, 21\penalty0 (1):\penalty0 257--270, 1988.

\bibitem[Geiping et~al.(2024)Geiping, Stein, Shu, Saifullah, Wen, and Goldstein]{geiping2024coercingllmsrevealalmost}
Geiping, J., Stein, A., Shu, M., Saifullah, K., Wen, Y., and Goldstein, T.
\newblock Coercing llms to do and reveal (almost) anything, 2024.
\newblock URL \url{https://arxiv.org/abs/2402.14020}.

\bibitem[Grunwald \& Vit{\'a}nyi(2004)Grunwald and Vit{\'a}nyi]{grunwald2004shannon}
Grunwald, P. and Vit{\'a}nyi, P.
\newblock Shannon information and kolmogorov complexity.
\newblock \emph{arXiv preprint cs/0410002}, 2004.

\bibitem[Grunwald \& Vitanyi(2004)Grunwald and Vitanyi]{grunwald2004shannoninformationkolmogorovcomplexity}
Grunwald, P. and Vitanyi, P.
\newblock Shannon information and kolmogorov complexity, 2004.
\newblock URL \url{https://arxiv.org/abs/cs/0410002}.

\bibitem[Haghifam et~al.(2020)Haghifam, Negrea, Khisti, Roy, and Dziugaite]{haghifam2020sharpened}
Haghifam, M., Negrea, J., Khisti, A., Roy, D.~M., and Dziugaite, G.~K.
\newblock Sharpened generalization bounds based on conditional mutual information and an application to noisy, iterative algorithms.
\newblock \emph{Advances in Neural Information Processing Systems}, 33:\penalty0 9925--9935, 2020.

\bibitem[Heinzerling \& Inui(2021)Heinzerling and Inui]{heinzerling2021languagemodelsknowledgebases}
Heinzerling, B. and Inui, K.
\newblock Language models as knowledge bases: On entity representations, storage capacity, and paraphrased queries, 2021.
\newblock URL \url{https://arxiv.org/abs/2008.09036}.

\bibitem[Jayaraman \& Evans(2022)Jayaraman and Evans]{jayaraman2022attribute}
Jayaraman, B. and Evans, D.
\newblock Are attribute inference attacks just imputation?
\newblock In \emph{Proceedings of the 2022 ACM SIGSAC Conference on Computer and Communications Security}, pp.\  1569--1582, 2022.

\bibitem[Kajitsuka \& Sato(2024)Kajitsuka and Sato]{kajitsuka2024optimalmemorizationcapacitytransformers}
Kajitsuka, T. and Sato, I.
\newblock Optimal memorization capacity of transformers, 2024.
\newblock URL \url{https://arxiv.org/abs/2409.17677}.

\bibitem[Kaplan et~al.(2020)Kaplan, McCandlish, Henighan, Brown, Chess, Child, Gray, Radford, Wu, and Amodei]{kaplan2020scalinglawsneurallanguage}
Kaplan, J., McCandlish, S., Henighan, T., Brown, T.~B., Chess, B., Child, R., Gray, S., Radford, A., Wu, J., and Amodei, D.
\newblock Scaling laws for neural language models, 2020.
\newblock URL \url{https://arxiv.org/abs/2001.08361}.

\bibitem[Kolmogorov(1963)]{kolmogorov1963tables}
Kolmogorov, A.~N.
\newblock On tables of random numbers.
\newblock \emph{Sankhyā: The Indian Journal of Statistics, Series A}, 25\penalty0 (4):\penalty0 369--376, 1963.
\newblock URL \url{http://www.jstor.org/stable/25049284}.
\newblock Accessed: 21/12/2010 15:32.

\bibitem[Kolmogorov(1965)]{kolmogorov1965three}
Kolmogorov, A.~N.
\newblock Three approaches to the quantitative definition of information.
\newblock \emph{Problems of Information Transmission}, 1\penalty0 (1):\penalty0 1--7, 1965.

\bibitem[Lee et~al.(2022)Lee, Ippolito, Nystrom, Zhang, Eck, Callison-Burch, and Carlini]{lee2022deduplicatingtrainingdatamakes}
Lee, K., Ippolito, D., Nystrom, A., Zhang, C., Eck, D., Callison-Burch, C., and Carlini, N.
\newblock Deduplicating training data makes language models better, 2022.
\newblock URL \url{https://arxiv.org/abs/2107.06499}.

\bibitem[Lu et~al.(2024)Lu, Li, Cheng, Ding, Huang, and Qiu]{lu2024scalinglawsfactmemorization}
Lu, X., Li, X., Cheng, Q., Ding, K., Huang, X., and Qiu, X.
\newblock Scaling laws for fact memorization of large language models, 2024.
\newblock URL \url{https://arxiv.org/abs/2406.15720}.

\bibitem[Mahdavi et~al.(2024)Mahdavi, Liao, and Thrampoulidis]{mahdavi2024memorizationcapacitymultiheadattention}
Mahdavi, S., Liao, R., and Thrampoulidis, C.
\newblock Memorization capacity of multi-head attention in transformers, 2024.
\newblock URL \url{https://arxiv.org/abs/2306.02010}.

\bibitem[Maini et~al.(2024)Maini, Jia, Papernot, and Dziedzic]{maini2024llmdatasetinferencedid}
Maini, P., Jia, H., Papernot, N., and Dziedzic, A.
\newblock Llm dataset inference: Did you train on my dataset?, 2024.
\newblock URL \url{https://arxiv.org/abs/2406.06443}.

\bibitem[Mireshghallah et~al.(2022)Mireshghallah, Uniyal, Wang, Evans, and Berg-Kirkpatrick]{mireshghallah2022memorizationnlpfinetuningmethods}
Mireshghallah, F., Uniyal, A., Wang, T., Evans, D., and Berg-Kirkpatrick, T.
\newblock Memorization in nlp fine-tuning methods, 2022.
\newblock URL \url{https://arxiv.org/abs/2205.12506}.

\bibitem[Nakkiran et~al.(2019)Nakkiran, Kaplun, Bansal, Yang, Barak, and Sutskever]{nakkiran2019deepdoubledescentbigger}
Nakkiran, P., Kaplun, G., Bansal, Y., Yang, T., Barak, B., and Sutskever, I.
\newblock Deep double descent: Where bigger models and more data hurt, 2019.
\newblock URL \url{https://arxiv.org/abs/1912.02292}.

\bibitem[Nasr et~al.(2023)Nasr, Carlini, Hayase, Jagielski, Cooper, Ippolito, Choquette-Choo, Wallace, Tramèr, and Lee]{nasr2023scalableextractiontrainingdata}
Nasr, M., Carlini, N., Hayase, J., Jagielski, M., Cooper, A.~F., Ippolito, D., Choquette-Choo, C.~A., Wallace, E., Tramèr, F., and Lee, K.
\newblock Scalable extraction of training data from (production) language models, 2023.
\newblock URL \url{https://arxiv.org/abs/2311.17035}.

\bibitem[Penedo et~al.(2024)Penedo, Kydlíček, allal, Lozhkov, Mitchell, Raffel, Werra, and Wolf]{penedo2024fineweb}
Penedo, G., Kydlíček, H., allal, L.~B., Lozhkov, A., Mitchell, M., Raffel, C., Werra, L.~V., and Wolf, T.
\newblock The fineweb datasets: Decanting the web for the finest text data at scale, 2024.
\newblock URL \url{https://arxiv.org/abs/2406.17557}.

\bibitem[Prashanth et~al.(2024)Prashanth, Deng, O'Brien, V, Khan, Borkar, Choquette-Choo, Fuehne, Biderman, Ke, Lee, and Saphra]{prashanth2024recitereconstructrecollect}
Prashanth, U.~S., Deng, A., O'Brien, K., V, J.~S., Khan, M.~A., Borkar, J., Choquette-Choo, C.~A., Fuehne, J.~R., Biderman, S., Ke, T., Lee, K., and Saphra, N.
\newblock Recite, reconstruct, recollect: Memorization in lms as a multifaceted phenomenon, 2024.
\newblock URL \url{https://arxiv.org/abs/2406.17746}.

\bibitem[Radford et~al.(2019)Radford, Wu, Child, Luan, Amodei, and Sutskever]{radford2019language}
Radford, A., Wu, J., Child, R., Luan, D., Amodei, D., and Sutskever, I.
\newblock Language models are unsupervised multitask learners.
\newblock 2019.

\bibitem[Roberts et~al.(2020)Roberts, Raffel, and Shazeer]{roberts2020knowledgepackparameterslanguage}
Roberts, A., Raffel, C., and Shazeer, N.
\newblock How much knowledge can you pack into the parameters of a language model?, 2020.
\newblock URL \url{https://arxiv.org/abs/2002.08910}.

\bibitem[Sablayrolles et~al.(2019)Sablayrolles, Douze, Ollivier, Schmid, and Jégou]{sablayrolles2019whiteboxvsblackboxbayes}
Sablayrolles, A., Douze, M., Ollivier, Y., Schmid, C., and Jégou, H.
\newblock White-box vs black-box: Bayes optimal strategies for membership inference, 2019.
\newblock URL \url{https://arxiv.org/abs/1908.11229}.

\bibitem[Schwarzschild et~al.(2024)Schwarzschild, Feng, Maini, Lipton, and Kolter]{schwarzschild2024rethinkingllmmemorizationlens}
Schwarzschild, A., Feng, Z., Maini, P., Lipton, Z.~C., and Kolter, J.~Z.
\newblock Rethinking llm memorization through the lens of adversarial compression, 2024.
\newblock URL \url{https://arxiv.org/abs/2404.15146}.

\bibitem[Shannon(1948)]{shannon1948noisychannel}
Shannon, C.~E.
\newblock \emph{A Mathematical Theory of Communication}.
\newblock University of Illinois Press, 1948.
\newblock Reprint in 1998.

\bibitem[Shannon(1950)]{shannon1950englishentropy}
Shannon, C.~E.
\newblock Prediction and entropy of printed english, Sept 1950.

\bibitem[Shokri et~al.(2017)Shokri, Stronati, Song, and Shmatikov]{shokri2017membership}
Shokri, R., Stronati, M., Song, C., and Shmatikov, V.
\newblock Membership inference attacks against machine learning models.
\newblock In \emph{2017 IEEE symposium on security and privacy (SP)}, pp.\  3--18. IEEE, 2017.

\bibitem[Shwartz-Ziv et~al.(2024)Shwartz-Ziv, Goldblum, Bansal, Bruss, LeCun, and Wilson]{shwartzziv2024justflexibleneuralnetworks}
Shwartz-Ziv, R., Goldblum, M., Bansal, A., Bruss, C.~B., LeCun, Y., and Wilson, A.~G.
\newblock Just how flexible are neural networks in practice?, 2024.
\newblock URL \url{https://arxiv.org/abs/2406.11463}.

\bibitem[Steinke \& Zakynthinou(2020)Steinke and Zakynthinou]{steinke2020reasoning}
Steinke, T. and Zakynthinou, L.
\newblock Reasoning about generalization via conditional mutual information.
\newblock In \emph{Conference on Learning Theory}, pp.\  3437--3452. PMLR, 2020.

\bibitem[Tänzer et~al.(2022)Tänzer, Ruder, and Rei]{tanzer2022memorisationversusgeneralisationpretrained}
Tänzer, M., Ruder, S., and Rei, M.
\newblock Memorisation versus generalisation in pre-trained language models, 2022.
\newblock URL \url{https://arxiv.org/abs/2105.00828}.

\bibitem[Xia et~al.(2023)Xia, Artetxe, Zhou, Lin, Pasunuru, Chen, Zettlemoyer, and Stoyanov]{xia2023trainingtrajectorieslanguagemodels}
Xia, M., Artetxe, M., Zhou, C., Lin, X.~V., Pasunuru, R., Chen, D., Zettlemoyer, L., and Stoyanov, V.
\newblock Training trajectories of language models across scales, 2023.
\newblock URL \url{https://arxiv.org/abs/2212.09803}.

\bibitem[Xu et~al.(2020)Xu, Zhao, Song, Stewart, and Ermon]{xu2020vinformation}
Xu, Y., Zhao, S., Song, J., Stewart, R., and Ermon, S.
\newblock A theory of usable information under computational constraints, 2020.
\newblock URL \url{https://arxiv.org/abs/2002.10689}.

\bibitem[Yeom et~al.(2018)Yeom, Giacomelli, Fredrikson, and Jha]{yeom2018privacyriskmachinelearning}
Yeom, S., Giacomelli, I., Fredrikson, M., and Jha, S.
\newblock Privacy risk in machine learning: Analyzing the connection to overfitting, 2018.
\newblock URL \url{https://arxiv.org/abs/1709.01604}.

\bibitem[Yun et~al.(2019)Yun, Sra, and Jadbabaie]{yun2019smallrelunetworkspowerful}
Yun, C., Sra, S., and Jadbabaie, A.
\newblock Small relu networks are powerful memorizers: a tight analysis of memorization capacity, 2019.
\newblock URL \url{https://arxiv.org/abs/1810.07770}.

\bibitem[Zhang et~al.(2017)Zhang, Bengio, Hardt, Recht, and Vinyals]{zhang2017understandingdeeplearningrequires}
Zhang, C., Bengio, S., Hardt, M., Recht, B., and Vinyals, O.
\newblock Understanding deep learning requires rethinking generalization, 2017.
\newblock URL \url{https://arxiv.org/abs/1611.03530}.

\bibitem[Zhang et~al.(2023)Zhang, Ippolito, Lee, Jagielski, Tramèr, and Carlini]{zhang2023counterfactualmemorizationneurallanguage}
Zhang, C., Ippolito, D., Lee, K., Jagielski, M., Tramèr, F., and Carlini, N.
\newblock Counterfactual memorization in neural language models, 2023.
\newblock URL \url{https://arxiv.org/abs/2112.12938}.

\end{thebibliography}
\bibliographystyle{neurips_2025}

\appendix

\section{Appendix}

\subsection{How reliable are our linear estimates of capacity?}

\begin{table*}[t]
    \small
    \begin{minipage}{0.4\textwidth}
    \centering
    \begin{tabular}{rrrrr}
        \toprule
        $S$ & Params. & Memorized & Expected & Error \\
        \midrule
         4   & $6.59 \times 10^5$ & $1.73 \times 10^5$  & $1.80 \times 10^5$  & 4.19 \\
         8   & $6.60 \times 10^5$ & $3.54 \times 10^5$  & $3.60 \times 10^5$  & 1.80 \\
         16  & $6.61 \times 10^5$ & $7.15 \times 10^5$  & $7.21 \times 10^5$  & 0.84 \\
         32  & $6.63 \times 10^5$ & $1.44 \times 10^6$ & $1.44 \times 10^6$ & 0.41 \\
         64  & $6.67 \times 10^5$ & $2.29 \times 10^6$ & $2.36 \times 10^6$ & 2.97 \\
         128 & $6.75 \times 10^5$ & $2.36 \times 10^6$ & $2.39 \times 10^6$ & 1.24 \\
         256 & $6.92 \times 10^5$ & $2.44 \times 10^6$ & $2.45 \times 10^6$ & 0.44 \\
        \bottomrule
    \end{tabular}
    \caption{Model capacity estimates across sequence length $S$, along with error (\%).}
    \label{tab:seq-len-estimates}
    \end{minipage}
    \hspace{1.6cm}
    \begin{minipage}{0.4\textwidth}
    \centering
    \begin{tabular}{rrrrr}
        \toprule
        $V$ & Params. & Memorized & Expected & Error \\
        \midrule
        128 & $4.21 \times 10^5$ & $1.49 \times 10^6$ & $1.49 \times 10^6$ & 0.36 \\
        512 & $4.71 \times 10^5$ & $1.71 \times 10^6$ & $1.67 \times 10^6$ & 2.78 \\
        1024 & $5.36 \times 10^5$ & $1.95 \times 10^6$ & $1.90 \times 10^6$ & 2.70 \\
        2048 & $6.67 \times 10^5$ & $2.39 \times 10^6$ & $2.36 \times 10^6$ & 1.11 \\
        4096 & $9.29 \times 10^5$ & $3.13 \times 10^6$ & $3.15 \times 10^6$ & 0.47 \\
        \bottomrule

    \end{tabular}
    \caption{Model capacity estimates across vocab size $V$, along with error (\%).}
    \label{tab:vocab-size-estimates}
    \end{minipage}
\end{table*}

\begin{figure*}[t]
    \centering
    \begin{minipage}{.47\textwidth}
        \centering
        \includegraphics[width=\textwidth]{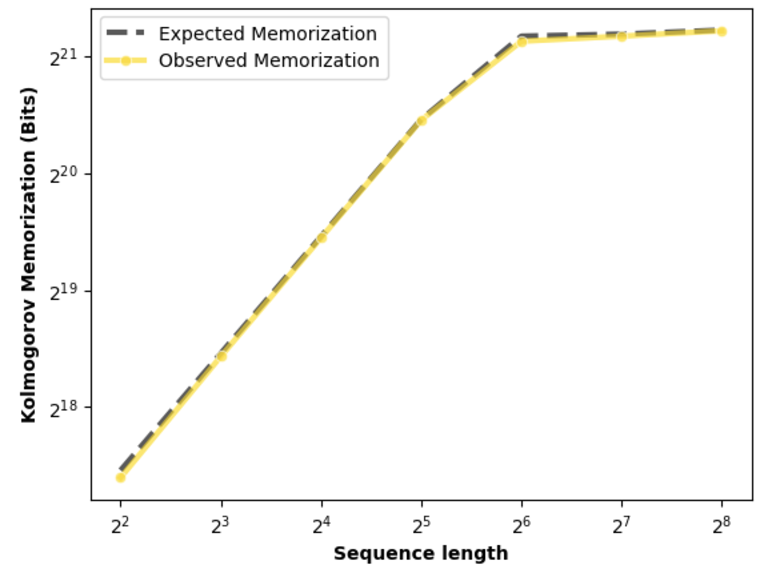}
        \caption{Model memorization across sequence lengths for a fixed-length dataset. Our predictions of total memorization are accurate, with an average error rate of 1.7\%.}
        \label{fig:synth-seqlen}
    \end{minipage}%
  \hspace{.5cm} 
    \begin{minipage}{.47\textwidth}
        \centering
        \includegraphics[width=\textwidth]{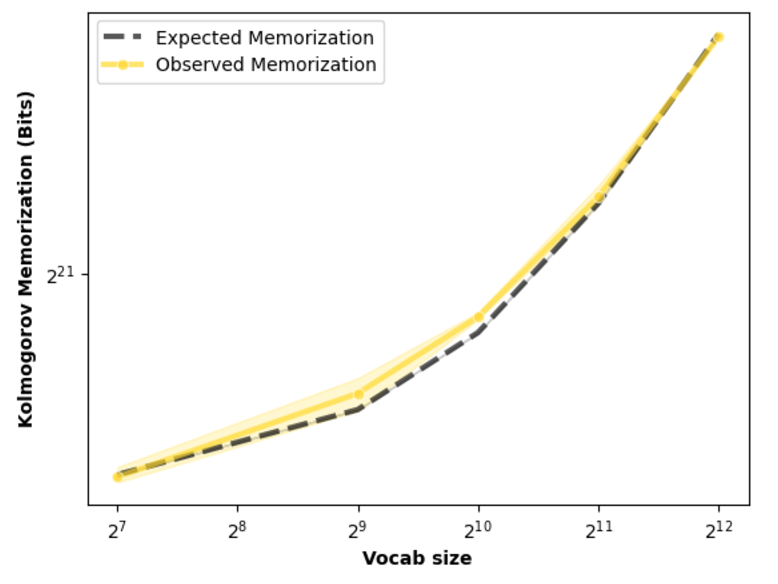}
        \caption{Model memorization across vocabulary size for a fixed-length dataset. Our predictions of total memorization are accurate, with an average error rate of 1.8\%. Note that, we do not observe a capacity plateau, since increasing $V$ also increases parameters.}
        \label{fig:synth-vocab}
    \end{minipage}
\end{figure*}

Instead of scaling the number of examples in a dataset, we scale model 
sequence length to adjust the size of a dataset. We use the following measurement for expected memorization of a model:

\[
\text{mem}(X,L(X)) \approx \min(capacity(L), H(X))
\]

we substitute our previous estimate of $\alpha=3.642$ and ensure to adjust the parameter count for increases due to resizing the model's embedding matrices. We fix the number of training samples to $4096$ and train a model with $2$ layers and a hidden size of $128$. Results are illustrated in Figure \ref{fig:synth-seqlen} and Table \ref{tab:seq-len-estimates}. Our predictions of total memorization are accurate, with an average error rate of 1.7\% while scaling $S$ and 1.8\% when scaling $V$.

\subsection{Additional memorization results}

Our findings indicate that memorization of text data neatly plateaus near the model capacity just as in the synthetic data case. When the dataset size increases by a factor of $N$, the model divides its memorization between datapoints by an equal amount; the sum of memorization is measured to be constant, presumably at the upper bound of the model's capacity.

When the dataset is small enough for each model to fit – that is, below the capacity of the smallest model – we observe very similar performance between the models. For larger data sizes we notice an interesting trend: \unintendedMemorizationName{} increases with dataset size for to a point, presumably as a model fills its capacity with the available information, and then decreases, as the model replaces sample-level information with more useful, generalizable knowledge. A given model generalizes the most (and memorizes the least information about any individual sample) when the dataset is maximally large.

\begin{figure*}[t]
    \centering
    \begin{minipage}[t]{0.47\textwidth}
        \centering
        \includegraphics[width=\textwidth]{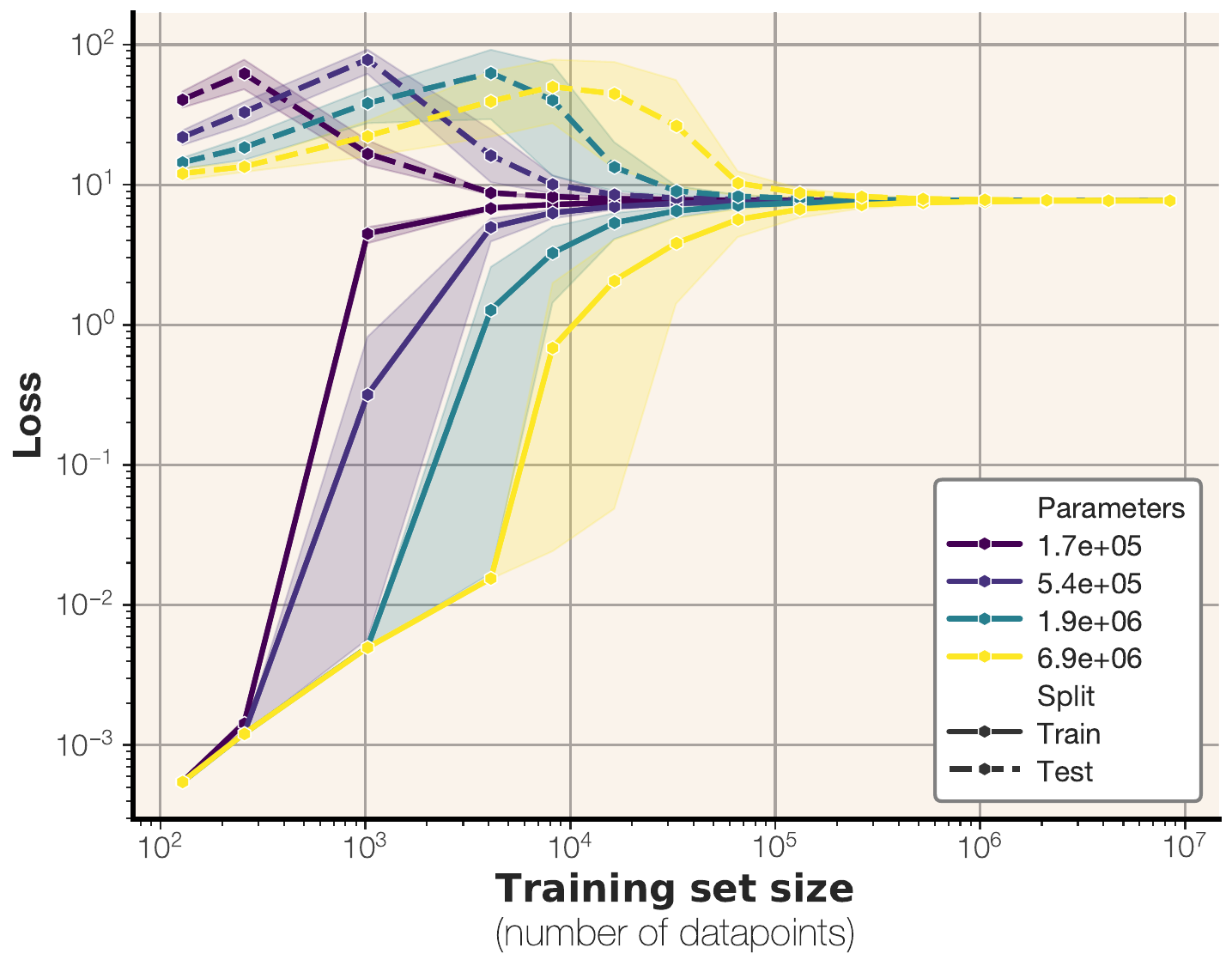}
        \caption{Train and test losses for different-sized language models trained on synthetic data.}
        \label{fig:synth_loss}
    \end{minipage}
    \hfill
    \begin{minipage}[t]{0.47\textwidth}
        \centering
        \includegraphics[width=\textwidth]{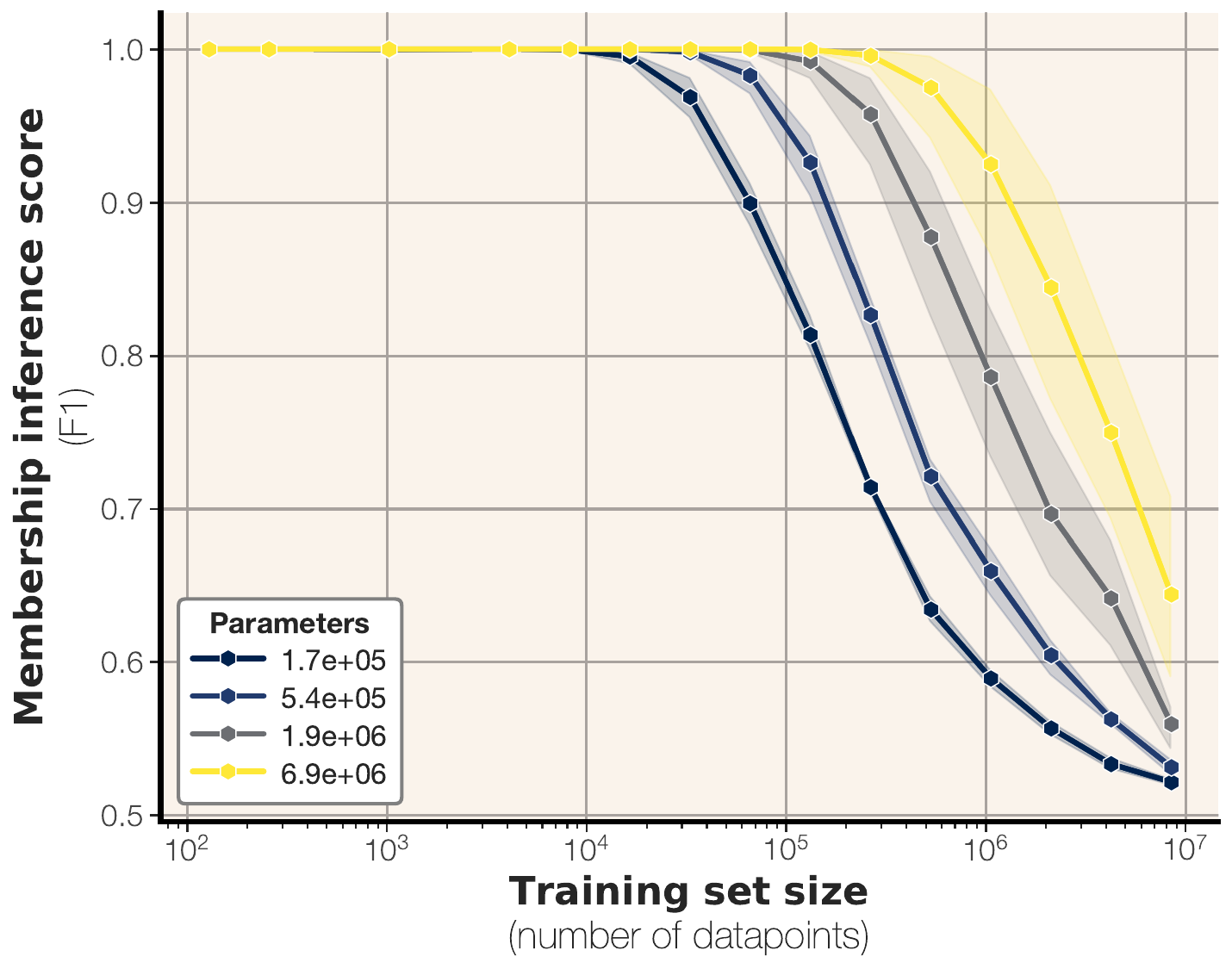}
        \caption{Membership inference attack performance decreases with dataset scale. In the case of uniform synthetic data, membership inference performance never falls below $0.54$.}
        \label{fig:model_capacity_membership}
    \end{minipage}
\end{figure*}

\subsection{Comparison of distributions memorized}

\begin{figure*}[t]
    \begin{minipage}[t]{0.47\textwidth}
        \centering
        \includegraphics[width=\textwidth]{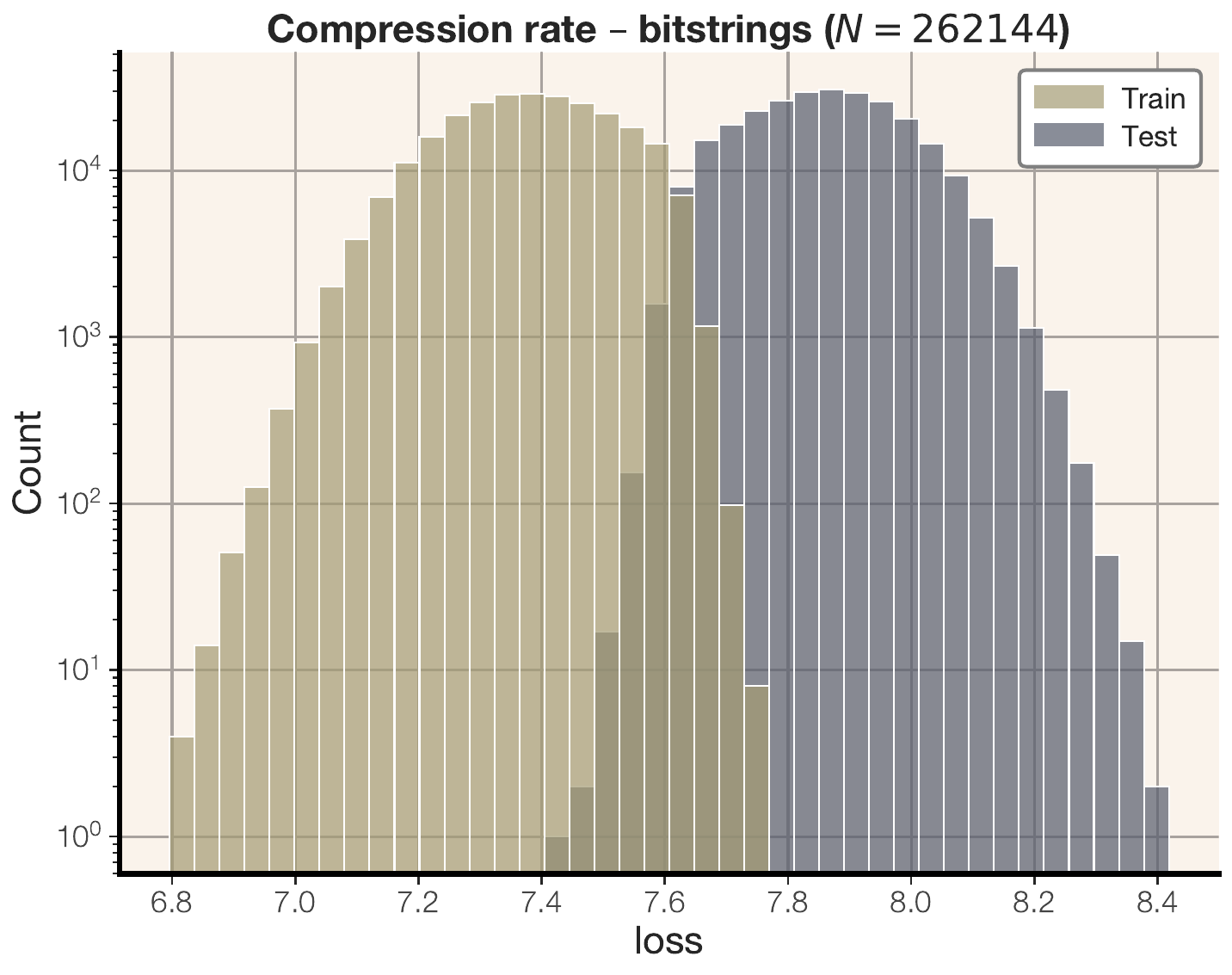}
    \end{minipage}
    \hspace{0.02\textwidth}%
    \begin{minipage}[t]{0.47\textwidth}
        \centering
        \includegraphics[width=\textwidth]{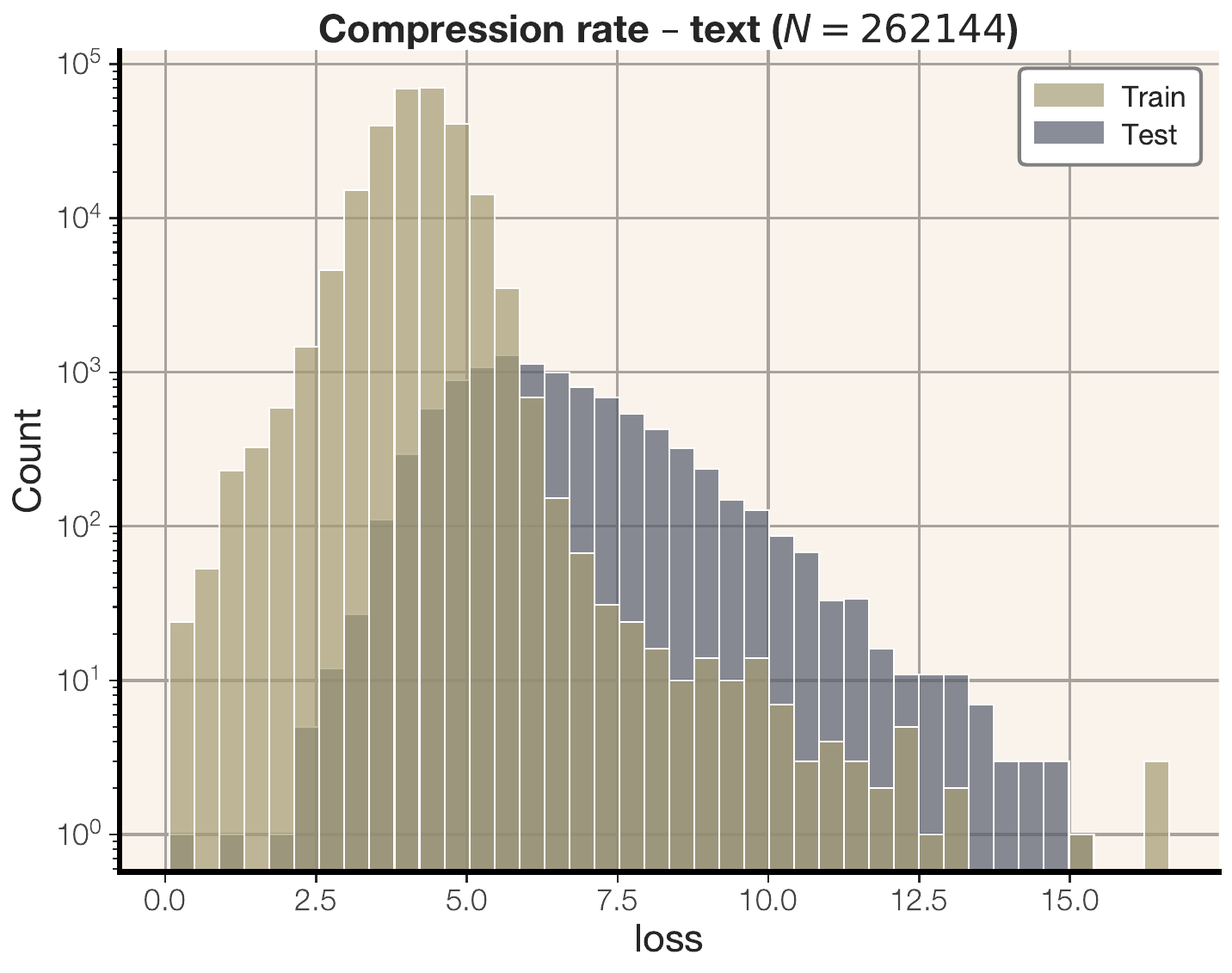}
    \end{minipage}
    \caption{Distribution of compression rates for equal-sized transformers ($n_\text{layer} = 4$, $d_\text{model}=128$) trained on $2^{14}$ sequences of equal-length random bitstrings (left) and text (right).}
    \label{fig:hist-synth-text}
\end{figure*}

\paragraph{Distribution-level analysis.} Text sequences have very different properties than uniform synthetic bitstrings. We explore how two models of equal capacity spread their memorization across datapoints. We plot a histogram (Figure \ref{fig:hist-synth-text}) of train and test compression rates of training data from both synthetic random bitstrings and text. Random training data follows a very normal distribution with a small amount of overlap between train and test compression rates. Text loss is lower on average but more spread out, with low loss on some training points and a long tail of higher losses. There is much more overlap between the train and test loss distributions, which explains why membership inference is more difficult for text data.

\paragraph{Which datapoints are most memorized?} Our distribution-level analysis indicates that unlike in the random-bitstring case, models trained on a large amount of text are able to memorize a small number of datapoints. Prior work has indicated that a large amount of this memorization can be due to duplicated training points
\citep{lee2022deduplicatingtrainingdatamakes} but our dataset is fully deduplicated so this cannot be an explanation in our case.

\begin{figure*}
\centering
    \begin{minipage}[t]{0.47\textwidth}
        \centering
        \includegraphics[width=\textwidth]{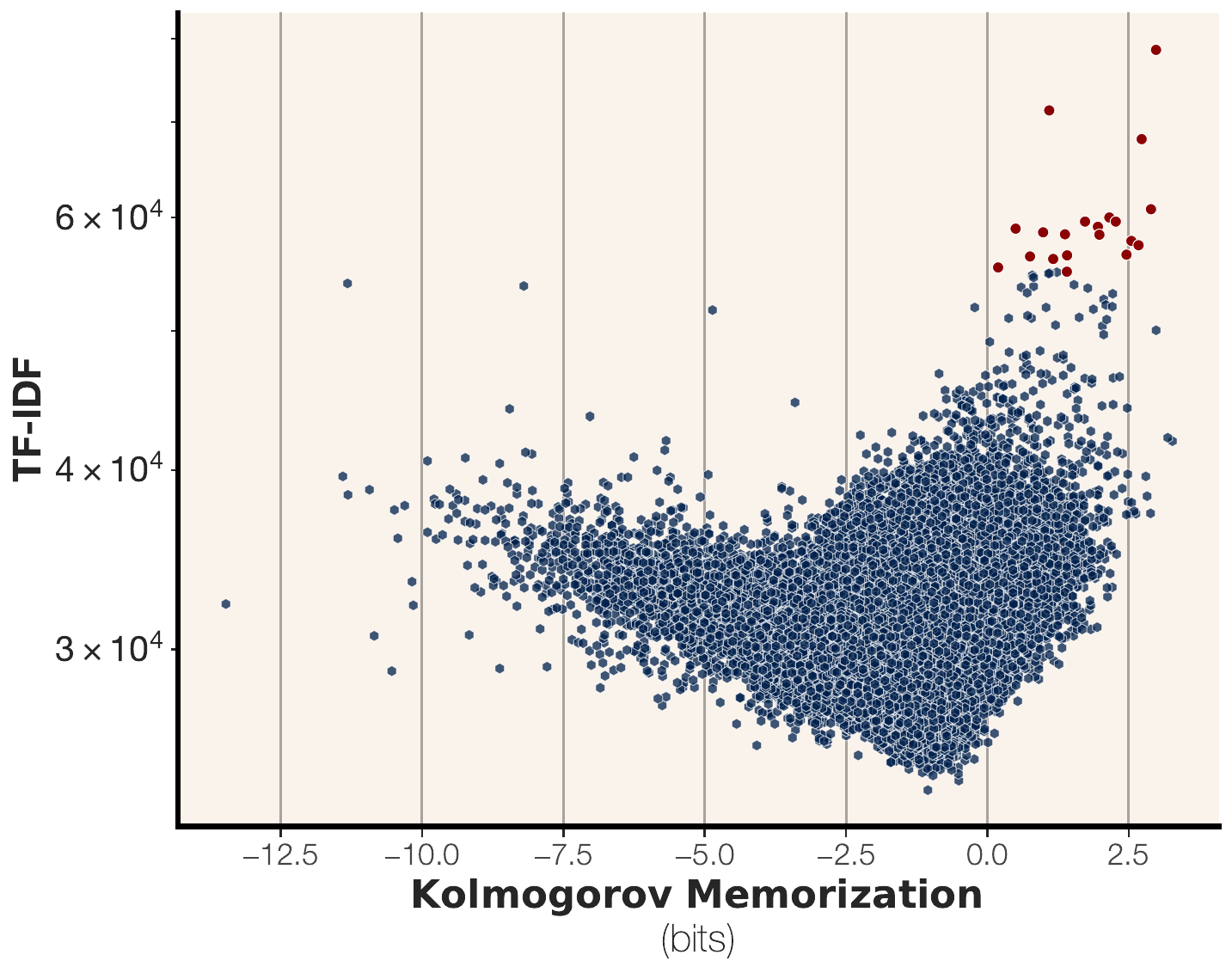}
    \end{minipage}
    \caption{\unintendedMemorizationNameCapital{}{} vs. TF-IDF for all training points of a $20M$ param model trained past its capacity on  $2^{16}$ sequences of English text. The training documents with rarest words are typically the most memorized.}
    \label{fig:text-mem-tfidf}
\end{figure*}
To quantitatively evaluate the number of rare words per document, we measure the TF-IDF of each training document, plotted vs. \unintendedMemorizationName{} in Figure \ref{fig:text-mem-tfidf}. We use the following equation for TF-IDF:

\[
\text{TF-IDF}(d; \mathcal{D}) = \frac{1}{|d|}\sum_{w \in d} \log \frac{|D|}{tf(w, \mathcal{D})}
\]

where $tf(d, \mathcal{D})$ indicates the total number of times word $w$ appears in dataset $\mathcal{D}$. Intuitively, a higher TF-IDF score for document $d$ indicates that $d$ contains more words that are rare in $\mathcal{D}$.

We clearly observe for samples with positive \unintendedMemorizationName{}  there is a strong correlation between trainset TF-IDF and memorization: examples with more rare words are more memorized. In particular, the sample with highest TF-IDF out of the whole training dataset (a sequence of Japanese words) has the third-highest measured memorization; even though this is just one out of $260,000$ training samples, the model can regurgitate the entire sequence given just a single token ({\begin{CJK*}{UTF8}{bkai}囚\end{CJK*}}). Out of the top twenty memorized sequences, all but three contain sequences of tokens from other languages (Japanese, Chinese, and Hebrew).

Manual analysis (Table \ref{tab:examples-table}) indicates that the most memorized datapoints have extremely rare tokens, typically ones not found in English. 

\begin{table*}[t]
    \centering
    \begin{tabular}{c}
        \includegraphics[width=\textwidth]{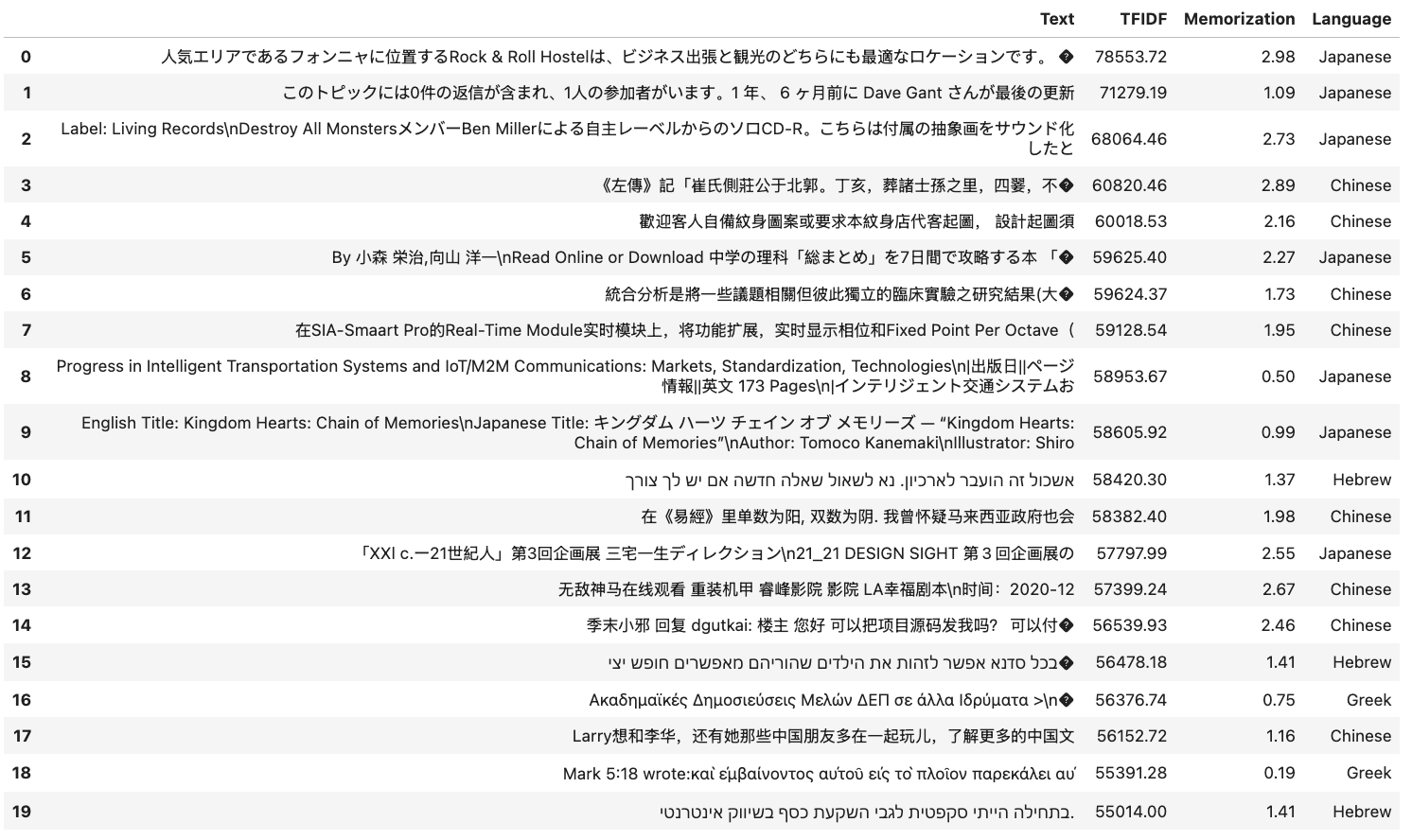} \\
    \end{tabular}
    \caption{Highest TF-IDF training examples from a $20M$ param model trained past its capacity on  $2^{16}$ sequences of English text. All of the highest TF-IDF examples are considered memorized, and contain text from non-English languages (Japanese, Chinese, Hebrew, and Greek).}
    \label{tab:examples-table}
\end{table*}

\subsection{Scaling law fit}
 
\begin{figure*}[t]
    \centering
    \includegraphics[width=.47\textwidth]{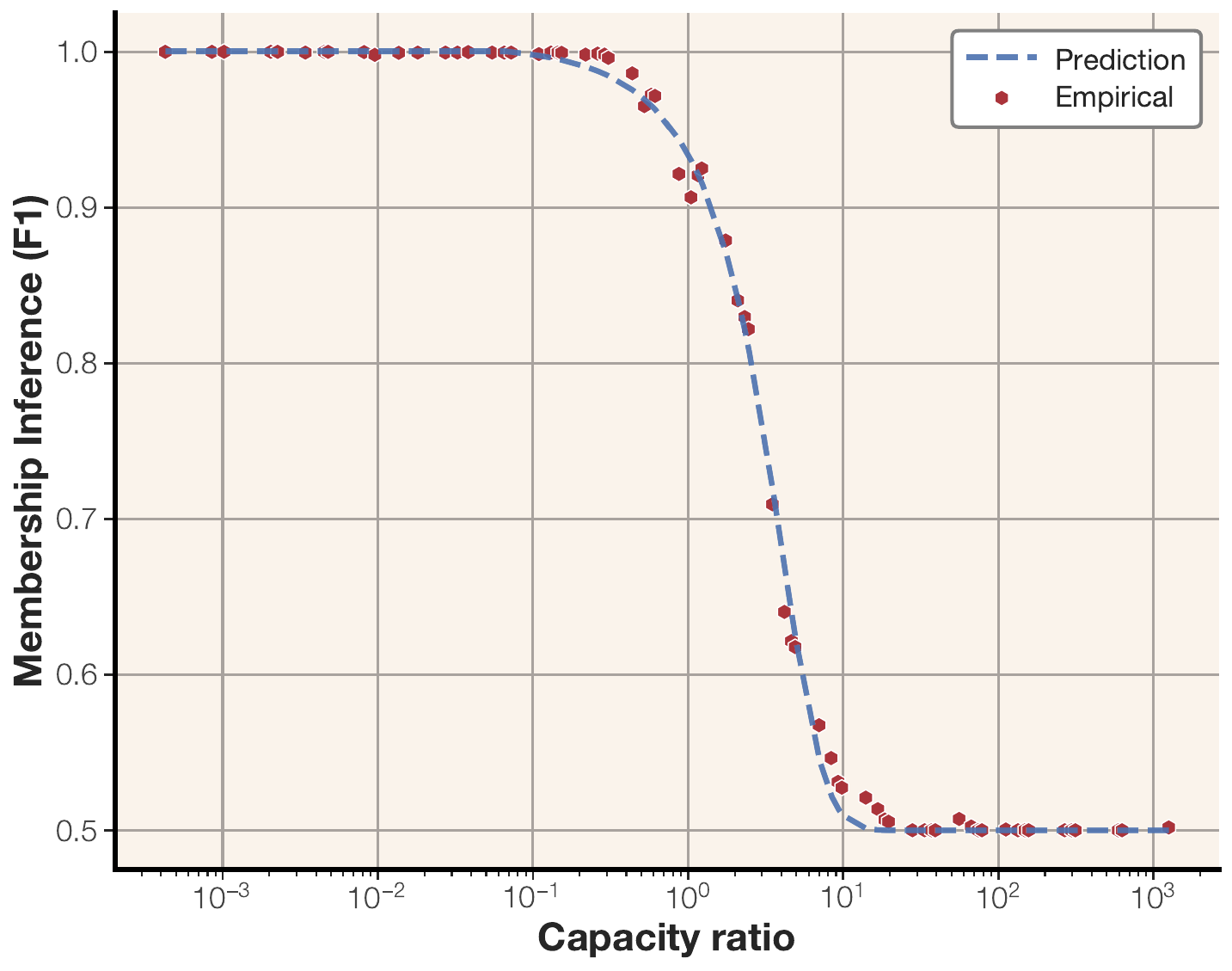}
    \caption{Our sigmoidal scaling law for membership inference fit to experimental data.}
    \label{fig:text-extract-eval}
\end{figure*}

Here we demonstrate the fit of our sigmoidal scaling law to experimental data. We show points in tokens-per-parameter vs. fit in Figure \ref{fig:text-extract-eval}. Although the sigmoidal function is slightly simplistic (the points do not perfectly fit the curve) our fit produces estimates within $1-2\%$ of observations.

\subsection{Proofs}
In the section we provide the proofs missing from the main body.
\subsection{Proof of Proposition \ref{prop:linearity_of_memorization}}
Here we prove Proposition \ref{prop:linearity_of_memorization}
\begin{proof} we have
\begin{align*}\text{mem}_U( X, \hat{\Theta},\Theta)&=I( X\mid \Theta, \hat{\Theta})\\
&= I((X_1 \mid \Theta, \dots, X_n \mid \Theta), \hat{\Theta}).
\end{align*}
And since the data is sampled i.i.d., all random variables in $\set{R_i=[X_i~\mid~\Theta]}_{i\in [n]}$ are independent. \footnote{Note that $X_i$ themselves are not independent because they are sampled by first sampling an underlying model $\Theta$. However, they are conditionally independent once the underlying model $\Theta$ is given.} So we have,
$$I((X_1 \mid \Theta, \dots, X_n \mid \Theta), \hat{\Theta}) \geq \sum_{i\in [n]} I(X_i \mid \Theta,\hat{\Theta})$$
which implies
$$\text{mem}_U(X, \hat{\Theta},\Theta) \geq \sum_{i\in[n]}{\text{mem}_U(X_i,\hat{\Theta}},\Theta).$$
On the other hand, we have

\begin{align*}
\text{mem}_U(X, \hat{\Theta},\Theta)&= I(X\mid \Theta, \hat{\Theta})\\ &=H(\hat{\Theta} - H(\hat\Theta \mid (X\mid \Theta))\\
&\leq H(\hat{\Theta})
\end{align*}
\end{proof}
\subsection{Proof of Proposition \ref{prop:kolmogorov_to_shannon}}
\begin{proof}
    We first state a Lemma about connection between algorithmic (kolmogorov) mutual information and mutual information.

    \begin{lemma}\label{lem:k_to_s}[Theorem 3.6 in \citet{grunwald2004shannon}] Assume $(X,Y)$ be a pair of joint random variables. Let $f$ be the density function, $f(x,y)=\Pr[(X,Y)=(x,y)].$ Then we have 
    
    \begin{align*}I(X,Y) - H_K(f)&\leq \Ex_{(x,y)\sim (X,Y)}[I_K(x,y)]\\
    &\leq I(X,Y) + 2H_K(f).
    \end{align*}
    \end{lemma}
    Now we use this lemma to prove the statement of the Proposition. Let $f$ be a the density function for the joint distribution $(X_i\mid \theta, \hat{\Theta})$. That is 
    $f_i(x_i, \hat{\theta}) = \Pr[X_i=x_i \mid \theta \text{~~and~~} \hat{\Theta}=\hat{\theta}]$. Note that this function is independent of $n$.   
    By definition we have
    $$\text{mem}_U(X_i, \hat{\Theta}, \theta) = I(X_i\mid \theta, \hat{\Theta}).$$
    Now using Lemma \ref{lem:k_to_s} we have 
    \begin{align*}I(X_i\mid \theta, \hat{\Theta}) -H_K(f)&\leq \Ex_{x_i\sim X_i\mid \theta}[I_K(x_i ,\hat{\theta})]\\
    &\leq I(X_i\mid \theta, \hat{\Theta})+ 2H_K(f).
    \end{align*}
    and this concludes the statement of Proposition by setting $\epsilon=2H_K(f)$
\end{proof}

\subsection{Limitations}

Our efforts to measure language model memorization come from a  line of recent research to discover whether models have analyzed certain texts, and if so, how much. However, our main experimental contributions relate to the practice of training and evaluating language models, including a new perspective on the phenomenon of grokking \citep{nakkiran2019deepdoubledescentbigger} and a new measurement of capacity. Our results are specific to the environment proposed and do not necessarily generalize to other datasets, architectures, or training setups.

\section{Discussion of other notions of memorization}
\label{app:memorization-related-work}
In this section we list multiple other notions of memorization and compare it with our definition. We specifically focus on why these notions do not satisfy all of our requirements.
\begin{itemize}
    \item {\textbf{Stability-based notions of memorization.}} There are notions of privacy and memorization that deal with ``stability'' of the training algorithm to small changes in the training set. Most notably, differential privacy \cite{dwork2006differential} considers the worst-cast drift of the model distribution when a single data point changes. Another notion of memorization in \citet{feldman2020does} is based on the change of the model prediction on a point $x$, when we add the labeled pair $(x,y)$ to the training set of a classification/regression model. Both of these notions are crucially relying on the learning algorithm and how it behaves. Moreover, the definition of differential privacy is not ideal for our case because it is a worst-case definition and cannot be applied at sample/model level. While the notion of memorization in \citet{feldman2020does} does not have this particular issue, it suffers from the fact that it only applies to classification models and mostly deals with the memorization of the association between the label ($y$) and input ($x$), and not the memorization of $x$ itself. These issues make these notions not ideal for our case. 
    \item {\textbf{Extraction-based memorization.}} There are multiple works in the literature \citep{carlini2019secretsharerevaluatingtesting,mireshghallah2022memorizationnlpfinetuningmethods,nasr2023scalableextractiontrainingdata,zhang2023counterfactualmemorizationneurallanguage,carlini2023quantifyingmemorizationneurallanguage,schwarzschild2024rethinkingllmmemorizationlens} that define memorization of samples in language models based on how easy it is to extract that sample. Specifically, when trying to understand the extent of memorization of a sample $x$ in a model $\theta$ they measure some notion of complexity for the task of eliciting the model to output $x$. Although these notions are great in that they only take a model $\theta$ and a sample $x$, they still do not account for generalization. Considering our running example of the following training sample: "What is $2^{100}$? (A: $1,267,650,600,228,229,401,496,703,205,376$)", this will be identified as highly memorized by almost all of the extraction based notions of memorization. Another issue with these definitions are that they are heavily dependent on the details of decoding algorithm. This is not ideal as we do not expect the memorization of a sample $x$ in a model $\theta$ to depend on the detailed parameters we use to generate samples using $\theta$.
    
    The work of \cite{schwarzschild2024rethinkingllmmemorizationlens} in this category is the closest to ours. This work which is based on prompt-optimization, optimizes a short  prompt $p$ to make the model elicit $x$, then it calls the sample $x$ memorized, if length of $p$ is less than $x$. Although this definition is close to our definition in using compression, it still does not account for generalization of the model. Moreover, it focuses on a specific way of compression through prompting. We posit that compression through prompting is an inferior compression scheme and can often lead to compression rates greater than 1.  
    \item {\textbf{Membership/attribute inference.}} Membership inference \cite{shokri2017membership} and attribute inference attacks \cite{jayaraman2022attribute} have been used for empirically measuring the privacy of machine learning algorithms. These notions which usually aim at approximating the stability notions of memorization are suffering from the same shortcomings. They rely heavily on the learning algorithm and the data distribution. Moreover, they fail at providing a sample level notion of memorization. For example, the obtained accuracy for membership inference attack is only meaningful in the population level. This is because various attack may have different true positives for membership, and the union of all these true positive across different attack may cover the entire training set, rendering it unusable as a sample level notion of memorization. 
    
    \item {\textbf{Data copying in generative models.}} There are some interesting notions of memorization designed specifically for generative modeling where a generative model may output a certain portion of training samples \citep{bhattacharjee2023data, carlini2023extractingtrainingdatadiffusion}. These notions are similar to extraction based definition of memorization but they are more lenient in that they only require extraction of part of the training data. However, they still suffer from the same challenges as of extraction based definitions.

\end{itemize}

\end{document}